\pdfoutput=1
\documentclass{article}
\usepackage{microtype}
\usepackage{graphicx}
\usepackage{booktabs}
\usepackage{tabularx}
\usepackage{hyperref}
\usepackage[preprint]{icml2026}
\usepackage{amsmath}
\usepackage{amssymb}
\usepackage{amsthm}
\theoremstyle{plain}
\newtheorem{proposition}{Proposition}
\theoremstyle{definition}
\newtheorem{definition}{Definition}
\newtheorem{assumption}{Assumption}
\newcommand{\nvfp}{NVFP4}

\icmltitlerunning{What the Loss Curve Misses: FP4 Pretraining and Rare Facts}

\begin{document}

\twocolumn[
  \icmltitle{What the Loss Curve Misses:\\FP4 Pretraining Slows the Acquisition of Rare Facts}
  \begin{icmlauthorlist}
    \icmlauthor{Chang Lu}{aff}
    \icmlauthor{Mengjiao Zhang}{aff}
  \end{icmlauthorlist}
  \icmlaffiliation{aff}{Independent Researcher}
  \icmlcorrespondingauthor{Chang Lu}{luchang.cs@gmail.com}
  \icmlkeywords{low-precision training, FP4, NVFP4, knowledge acquisition, language models}
  \vskip 0.3in
]
\printAffiliationsAndNotice{}
\hypersetup{pdfsubject={Preprint: low-precision pretraining and rare-fact acquisition}}

\begin{abstract}
Low-precision pretraining is usually judged by validation loss. We ask whether matching loss also means learning rare
facts as efficiently. We train transformers from scratch with emulated 4-bit matrix multiplications on synthetic
biographies with controlled exposure counts, on their own and mixed into web text. \nvfp{} needs 1.32$\times$ and
1.52$\times$ as many exposures as BF16 to match its recall while validation loss differs by only 0.5\% and 2\%, and
1.26$\times$ and 1.30$\times$ when both models are evaluated with BF16 matrix multiplications. Under our primary
readout, the exposure cost is about 1.2$\times$ at matched loss. Zero-mean gradient noise reproduces the cost of
quantizing only the backward pass, which barely moves the loss. In a controlled experiment at twice the \nvfp{}
error, noise fed only to Adam's second-moment estimate acts like a lower learning rate and its cost fades, whereas
noise in the update lowers its signal-to-noise ratio and carries about two thirds of the cost through the end of
training. Raising the learning rate to offset the smaller average step does not recover the cost, either for this
noise or for \nvfp{}. Gradient statistics of BF16 checkpoints rank the costs of formats, partial quantizations and noise controls
without fitting (Spearman 0.96). We suggest measuring
rare-fact acquisition alongside loss when evaluating low-precision recipes.
\end{abstract}

\begin{figure*}[!t]
\centering
\includegraphics[width=\textwidth]{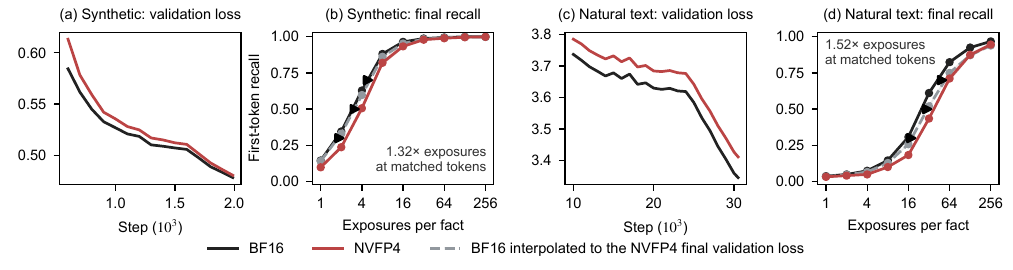}
\caption{BF16 (black) and \nvfp{} (red) training of the same 40M architecture, mean over six seeds per regime.
(a, c) Validation loss over the last 70\% of training. (b, d) Final recall against exposures per fact ($\pm$1
standard error over seeds), with BF16 recall interpolated to the \nvfp{} final validation loss in gray. Arrows mark
the horizontal shift of the mean curves at recall 0.3, 0.5 and 0.7.}
\label{fig:overview}
\end{figure*}

\section{Introduction}
Two training runs can produce nearly identical loss curves and still leave models that know different things. Train
the same transformer twice on the same data, once with BF16 matrix multiplications and once with NVIDIA's 4-bit
\nvfp{} format. The final validation losses differ by half a percent, yet the 4-bit model needs about a third more
exposures to learn the same facts (Figure~\ref{fig:overview}).

This gap matters because low-precision training is judged by loss. FP8 training is established at
scale~\citep{micikevicius2022fp8,peng2023fp8lm,fishman2025fp8,deepseek2024v3}, and 4-bit (FP4) recipes now approach
the loss and benchmark scores of higher
precision~\citep{wang2025fp4,castro2025quartet,chmiel2025fp4,tseng2025mxfp4,agrusa2026nvfp4,cao2026averis}. The
published \nvfp{} recipe, for example, trains a 12B model on 10T tokens close to an FP8
baseline~\citep{nvidia2025nvfp4}. Knowledge of the long tail, however, such as facts about less popular
entities~\citep{kandpal2023longtail,mallen2023popqa}, must be learned from few
exposures~\citep{chang2024acquire,allenzhu2024capacity}, and token-averaged loss gives such facts little weight. We
therefore ask whether low precision changes \emph{how many exposures a model needs to learn a fact}.

We answer by controlling exposure directly. We train transformers from scratch on synthetic
biographies~\citep{allenzhu2023physics31} whose subjects appear 1 to 256 times, on their own and mixed into
FineWeb-Edu~\citep{penedo2024fineweb}, with BF16 or emulated FP8, MXFP4 and \nvfp{} matrix multiplications. The
\emph{exposure multiplier}, the factor by which a low-precision model needs more exposures to reach the same recall,
condenses the gap into one number. \nvfp{} needs 1.32$\times$ as many exposures on synthetic data and 1.52$\times$ on
natural text at matched training tokens, and about 1.2$\times$ at matched validation loss. Reading both models with
BF16 matrix multiplications, which removes quantization error at evaluation time, leaves 1.26$\times$ and
1.30$\times$, so the deficit is not only an evaluation-time effect. Unlike the knowledge loss reported after
compression or the collapse of ternary training~\citep{jaiswal2023compressing,paudel2026linger}, this cost arises with
emulated production formats and shows up as slower learning without any collapse.

To locate the cost, we quantize the forward and backward passes separately. Quantizing only the backward pass leaves
the loss essentially unchanged but costs exposures, and adding zero-mean noise of the same size and shape to BF16
gradients does the same. Since such noise leaves the expected gradient unchanged, we look for its cost in how the
optimizer uses the gradient. Adam~\citep{kingma2015adam} divides each update by a running estimate of the gradient's
magnitude. Gradient noise inflates that estimate and so shrinks every step, as a lower learning rate would.
Noise also takes the place of signal in the update. Feeding controlled noise to only one of Adam's two moments
separates the two effects. Over our training runs the step-size cost fades, while the signal-to-noise cost persists
and carries about two thirds of the total. Raising the learning rate to offset the smaller average step does not
reduce the cost for the noise or for \nvfp{}, in tests whose criteria we fixed in advance. Full FP4 training also introduces
biased gradient error, so this decomposition accounts for the noise-driven part of its cost rather than all of it.
The step-size part of this picture, computed from gradient statistics of a BF16 checkpoint, already ranks the costs
of the formats, partial quantizations and noise controls we test (Spearman 0.96).

Remedies judged by loss can overstate how much factual acquisition recovers. Our adaptation of the published
\nvfp{} recipe closes 60\% of the loss gap but only 42\% of the exposure cost, and on natural text an estimated eighth
of it, with a wide interval. Exposure-controlled recall requires only an instrumented
slice of the corpus.

We summarize our contributions as follows.
\begin{itemize}\setlength{\itemsep}{1pt}\setlength{\parskip}{0pt}
\item \textbf{Measurement.} The exposure multiplier measures a cost that loss hides. \nvfp{} pretraining needs
1.2--1.5$\times$ as many exposures to reach the same recall at nearly equal loss, in every seed of the primary
comparisons (Sections~\ref{sec:main} and~\ref{sec:scale}).
\item \textbf{Mechanism.} In controlled interventions, zero-mean gradient noise suffices, and Adam turns it into a
step-size cost that fades and a signal-to-noise cost that persists over training and that a larger learning rate
does not remove (Sections~\ref{sec:model} and~\ref{sec:mech}).
\item \textbf{Practice.} Checkpoint gradient statistics rank costs without fitting, and tested remedies close the
loss gap faster than the exposure gap (Sections~\ref{sec:mech}
and~\ref{sec:recipe}).
\end{itemize}

\section{Measuring and Modelling the Cost}\label{sec:model}
\paragraph{The exposure multiplier.}
We compare a low-precision run with its BF16 reference through their recall curves $R(x)$, recall as a function of
$x=\log_2$ exposures, made nondecreasing by a running maximum. For a curve $R$ let $x_R(\ell)=\inf\{x: R(x)\ge\ell\}$
be the log-exposure at which it first reaches recall $\ell$.
\begin{definition}[Exposure multiplier]\label{def:mult}
For recall curves $R_{\mathrm{lo}}$ (low precision) and $R_{\mathrm{ref}}$ (reference), defined for all real $x$, and
a set of recall levels $L$,
let $L'\subseteq L$ contain the levels $\ell$ at which $x_{R_{\mathrm{lo}}}(\ell)$ and $x_{R_{\mathrm{ref}}}(\ell)$ are
both finite, and $\Delta_\ell=x_{R_{\mathrm{lo}}}(\ell)-x_{R_{\mathrm{ref}}}(\ell)$. The exposure
multiplier is $m=2^{\bar\Delta}$ with $\bar\Delta=|L'|^{-1}\sum_{\ell\in L'}\Delta_\ell$.
The multiplier is undefined if $L'$ is empty, and for a single level, $L=\{\ell\}$, it is $2^{\Delta_\ell}$.
\end{definition}
\begin{proposition}\label{prop:shift}
(i) If $R_{\mathrm{lo}}$ is $R_{\mathrm{ref}}$ delayed by a constant $\delta$ in $\log_2$ exposures,
$R_{\mathrm{lo}}(x)=R_{\mathrm{ref}}(x-\delta)$ for all $x$, then $\Delta_\ell=\delta$ for every $\ell\in L'$ and
$m=2^{\delta}$ whenever $L'$ is nonempty. (ii) For any strictly increasing $h$, replacing both curves by $h\circ R$ and
$L$ by $h(L)$ leaves the gap at each level $h(\ell)$ equal to $\Delta_\ell$, and hence $m$ unchanged.
\end{proposition}
Part (i) states what $m$ measures. If low precision delays learning by a constant factor $2^\delta$ in exposures, $m$
is that factor, whatever levels are used. Part (ii) states that a recall measure which is a strictly increasing
function of another yields the same multiplier for correspondingly transformed levels. These are properties of the
underlying curves, which finite-grid interpolation does not always preserve (Appendix~\ref{app:proof-shift}).

\begin{figure*}[!t]
\centering
\includegraphics[width=\textwidth]{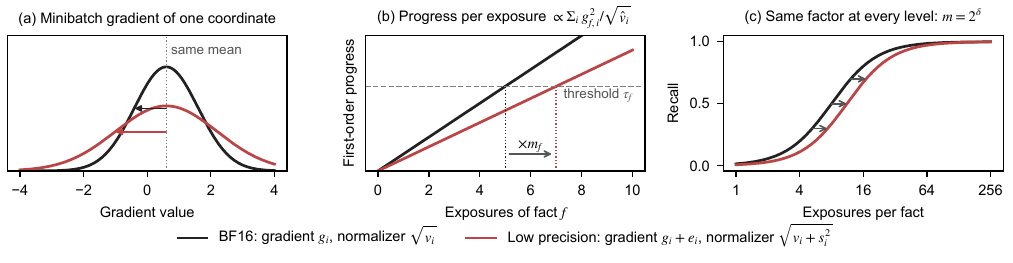}
\caption{The model of Section~\ref{sec:model} for one coordinate and one fact (illustrative shapes). (a) Zero-mean
gradient error widens the distribution of the minibatch gradient and raises Adam's normalizer from $\sqrt{v_i}$ to
$\sqrt{v_i+s_i^2}$. (b) Each exposure then adds less first-order progress, so the threshold $\tau_f$ is reached after
$m_f$ times as many exposures (Proposition~\ref{prop:adam}). (c) A common delay $m_f$ shifts the recall curve by
$\log_2 m_f$ at every level (Proposition~\ref{prop:shift}).}
\label{fig:schematic}
\end{figure*}

\paragraph{One exposure under Adam.}
We model AdamW's adaptive update~\citep{loshchilov2019adamw}, which divides the update of coordinate $i$ by
$\sqrt{\hat v_i}$, where $\hat v_i$ is Adam's running estimate of the second moment $v_i=\mathbb{E}[g_i^2]$ of the
minibatch gradient $g$. A
low-precision training configuration replaces $g$ by $g+e$, where $e$ is the error that quantized matrix
multiplications (GEMMs) add to the gradient, with per-coordinate second moment $s_i^2=\mathbb{E}[e_i^2]$. For a fact $f$, one attribute value of one
person (Section~\ref{sec:setup}), let $g_f$ be the gradient of the loss on its value tokens (the tokens of that value),
which one exposure contributes.
\begin{assumption}\label{ass:acc}
(a) Locally, the two configurations share fixed $g_f\ne0$, $v_i>0$ and learning rate $\eta$, and $s_i^2$ is also
fixed. The error
has zero mean given $g$, the error and the minibatch fluctuations of $g$ are uncorrelated across steps, and the
second-moment estimates equal their expectations.
Neglecting Adam's $\epsilon$, the normalizers are $\sqrt{v_i}$ and $\sqrt{v_i+s_i^2}$. (b) An exposure's momentum
contribution is integrated at these fixed normalizers. Recall occurs when accumulated first-order progress exceeds a
format-independent threshold $\tau_f$. Forgetting, curvature, weight decay and interactions between facts are omitted.
\end{assumption}
\begin{proposition}\label{prop:adam}
Let $\rho_i=s_i^2/v_i$. (i) Under Assumption~\ref{ass:acc}(a), an exposure of $f$ moves coordinate $i$ by
$-\eta\,g_{f,i}/\sqrt{v_i+s_i^2}$ in expectation instead of $-\eta\,g_{f,i}/\sqrt{v_i}$. Under
Assumption~\ref{ass:acc}, the ratio of the exposure counts needed to recall $f$ in the low-precision and reference
runs, treated as continuous, is
\begin{equation}\label{eq:mf}
m_f=\frac{\sum_i g_{f,i}^2/\sqrt{v_i}}{\sum_i g_{f,i}^2/\sqrt{v_i+s_i^2}}\;\ge 1
\end{equation}
for every threshold $\tau_f$. If $\rho_i=r$ for all $i$, then $m_f=\sqrt{1+r}$, and scaling the error by $c$ gives
$m_f(c)=\sqrt{1+c^2r}$. (ii) Under Assumption~\ref{ass:acc}(a), if $\mathbb{E}[g_i]^2\ll v_i$, the zero-mean part of the
update of coordinate $i$ keeps its variance to leading order, so the ratio of the update's mean to its standard
deviation falls by the factor $\sqrt{1+\rho_i}$. (iii) With the noisy gradient only in Adam's second moment, the
update of coordinate $i$ equals the reference update at learning rate $\eta/\sqrt{1+\rho_i}$. With it only in the first
moment, the expected update is unchanged and its variance grows by the factor $1+s_i^2/(v_i-\mathbb{E}[g_i]^2)$.
\end{proposition}
Part (i) is a step-size effect. Zero-mean error dilutes the expected progress of every exposure by $m_f$, which
depends only on where the error falls relative to the gradient noise. If the error matched the gradient noise on every
coordinate ($\rho_i=1$), each exposure would count as $1/\sqrt{2}$ of a reference exposure. Part (ii) is a
signal-to-noise effect. Where gradient noise dominates, Adam's normalized update keeps its size, so the error takes
the place of signal. Part (iii) gives two interventions that separate the effects (Section~\ref{sec:mech}). A change
of learning rate scales the expected update and its noise together, so it can restore the expected step of part (i)
but not the signal-to-noise ratio of part (ii). Under Assumption~\ref{ass:acc}, $m_f$ does not depend on $\tau_f$, so progress
towards every threshold is delayed alike (Figure~\ref{fig:schematic}). Proposition~\ref{prop:adam} yields four
predictions.
\begin{itemize}\setlength{\itemsep}{1pt}\setlength{\parskip}{0pt}
\item[(P1)] The single-level multiplier $2^{\Delta_\ell}$ is approximately constant across recall levels if $m_f$
varies little across facts. Zero-mean error does not guarantee this.
\item[(P2)] For small errors, scaling the error by $c$ gives $m_f(c)=1+c^2\bar\rho_f/2+O(c^4)$, where
$\bar\rho_f=\langle s^2/v\rangle_w$, and $\langle\cdot\rangle$ and $\langle\cdot\rangle_w$ denote the plain mean over
coordinates and the mean weighted by $w_i=g_{f,i}^2/\sqrt{v_i}$.
\item[(P3)] At equal total error power $\sum_i s_i^2$, an error with the same $s_i^2$ on every coordinate has a
larger coefficient $\bar\rho_f$ in P2 than one with $s_i^2\propto v_i$ when $\langle 1/v\rangle_w>1/\langle v\rangle$
(Appendix~\ref{app:proof-adam} gives sufficient conditions). Block-scaled rounding is closer to the proportional case, so
the ordering must be checked empirically.
\item[(P4)] Noise confined to the second moment acts as a per-coordinate learning-rate cut whose effect on a fact's
first-order progress equals a uniform cut by $m_f$. Where a uniform cut describes it well, its cost should match BF16
trained at the rate divided by $m_f$.
\end{itemize}
Measuring $v$, $s^2$ and $g_f$ on a BF16 checkpoint gives a prediction without fitting to exposure multipliers
(Section~\ref{sec:mech}). Biased gradient error also changes the expected update, which Eq.~\eqref{eq:mf} takes as
$g_{f,i}/\sqrt{v_i+s_i^2}$, and adds the cross term $2\mathbb{E}[g_i e_i]$ to its second moment. Applying
Eq.~\eqref{eq:mf} to such configurations tests it only as a diagnostic, outside the assumption.
Assumption~\ref{ass:acc}(b) prices only the expected step, so it implies that a larger learning rate always speeds
acquisition. In the model, an interior optimum such as the one we observe (Section~\ref{sec:scale}) arises from the
update noise of part (ii). A linearized model of one coordinate (Appendix~\ref{app:linear}) shows how the two effects
differ over time. A smaller step only slows the approach to the same time-averaged displacement of a repeatedly seen
fact, whereas update noise raises the stationary spread of the weights. The first suggests a delay that fades, and
the second a lasting penalty.

\paragraph{What validation loss sees.}
In the synthetic regime the validation stream consists of 2,000 freshly sampled biographies whose persons are drawn
in proportion to their exposures, as in the training stream.
\begin{proposition}\label{prop:loss}
Let facts at log-exposure $x$ have differentiable value-token loss $\mathcal{L}(x)$ under the reference and $\mathcal{L}(x-\delta)$
under low precision, for a constant delay $\delta\ge0$ in $\log_2$ exposures. Let a fraction $q(x)$ of facts have
log-exposure $x$, and the validation stream contain each fact in proportion to $2^x$. Then the excess value-token loss is
\begin{align*}
\Delta\mathcal{L}&=\frac{\sum_x q(x)\,2^x\,[\mathcal{L}(x-\delta)-\mathcal{L}(x)]}{\sum_x q(x)\,2^x}\\
&=\delta\,\mathbb{E}_{x\sim\pi}\big[-\mathcal{L}'(\xi_x)\big]
\end{align*}
with $\pi(x)\propto q(x)\,2^x$ and some $\xi_x\in[x-\delta,x]$. If recall is a fixed decreasing function of the
value-token loss, then $\Delta_\ell=\delta$ for every $\ell\in L'$ and $m=2^\delta$.
\end{proposition}
Loss thus weights the delay by the learning-curve slope under a distribution tilted towards frequent facts. If
those facts are already learned ($\mathcal{L}'\approx0$), their loss changes little despite a nonzero multiplier.
With nine equally populated bins, the most frequent
bin holds half of all value tokens and the two most frequent bins three quarters. Template words, which every
biography repeats, dilute the gap further.

\section{Setup}\label{sec:setup}
\paragraph{Facts with controlled exposure.}
Following the synthetic biography (bioS) construction~\citep{allenzhu2023physics31}, each of $N$ persons has six
attributes (birth date, birth city, university, major, employer, work city), and each attribute value of a person is
one fact. One exposure is one biography, with six sentences in random order, each from one of five templates. Persons
are split evenly into nine bins seen $2^0,\dots,2^8$ times. The $k$-th person goes to bin $k \bmod 9$, and names and
attribute values are drawn independently, so the bins are exchangeable. The facts and their bins are fixed across
seeds and configurations, and all exposures are shuffled into one stream. Recall is probed for 200 persons per bin
with a prompt per attribute, the first of its five training templates cut before the value. We report the accuracy
of the first value token (Appendix~\ref{app:example} shows a biography and its probes). Scoring exact match of the whole value instead moves every multiplier in Figure~\ref{fig:mult} by at most
0.07 (Appendix~\ref{app:robust}).

\paragraph{Regimes.}
\emph{Synthetic}: the stream consists of biographies only (word-level vocabulary). $N$ is set so that the run has a
target training budget (0.7--5.1 tokens per parameter). \emph{Natural}: biographies of 20,000 persons are inserted
as separate documents at random boundaries in FineWeb-Edu~\citep{penedo2024fineweb}. The resulting 1B-token mixture
contains 72M biography tokens (7.2\%) and uses a 16k byte-level BPE vocabulary. Validation loss is measured on
held-out FineWeb-Edu documents without inserted biographies.

\paragraph{Models and training.}
We use pre-norm transformers with RMSNorm~\citep{zhang2019rmsnorm}, rotary embeddings~\citep{su2021rope}, SwiGLU
MLPs~\citep{shazeer2020glu} and untied embeddings. The architectures are labeled 12M (6 blocks, width $d{=}384$,
10.6M non-embedding parameters), 40M (8 blocks, width 512, 25.7M) and 100M (12 blocks, width 768, 92.0M).
With the synthetic vocabulary, their total sizes are 12.9M, 28.7M and 96.6M, and the 40M architecture has 42.5M
parameters with the natural-text vocabulary. Tokens per parameter count all parameters~\citep{hoffmann2022chinchilla}.
We train with AdamW~\citep{loshchilov2019adamw} ($(\beta_1,\beta_2)=(0.9,0.95)$, weight decay 0.1 on matrices only,
gradients clipped to global norm 1), a peak learning rate of $10^{-3}$ and a warmup--stable--decay
schedule~\citep{hu2024minicpm,hagele2024wsd} with 2\% warmup and linear decay to zero over the last 20\% of steps.
The batch is $64\times512$ tokens at every size (100M uses four micro-batches per optimizer step). Seeds change the
initialization and the random draws of training (stochastic rounding, Hadamard signs, injected noise). The fact set
and the order of the training stream are the same in every run.

\paragraph{Low-precision configurations.}
All transformer-block linear layers quantize both operands of the forward, data-gradient and weight-gradient GEMMs
along the reduction dimension. Embeddings, the output head, attention score computations, normalization and optimizer
state stay in FP32. The BF16 reference rounds the operands of these GEMMs to BF16, and master weights remain FP32 in
every run. \nvfp{} uses E2M1 elements (2 exponent bits and 1 mantissa bit, with E4M3 and E5M2 named alike) in
16-element blocks with E4M3 scales under an FP32 tensor scale, stochastic rounding~\citep{gupta2015limited} in the
backward pass, and a random 16-point Hadamard transform (RHT)~\citep{xi2023int4,ashkboos2024quarot} on the
weight-gradient inputs. We call this configuration \emph{plain} \nvfp{} when comparing it with the recipe variants of
Section~\ref{sec:recipe}. MXFP4 (microscaling) uses E2M1 elements in 32-element blocks with power-of-two
scales~\citep{rouhani2023mx}. FP8 uses per-tensor E4M3 forward and E5M2 backward~\citep{micikevicius2022fp8}. GEMMs are emulated by quantizing operands exactly to the format and multiplying in
FP32 (TF32 disabled). This isolates operand rounding and scaling. We do not validate bitwise agreement with Blackwell
kernels or measure native FP4 throughput. All runs use one NVIDIA RTX 4090.

\paragraph{Computing the multiplier.}
The recall levels are $L=\{0.2,0.3,\dots,0.8\}$ (20--80\%), and recall is observed at $x=0,\dots,8$ and interpolated
linearly in $x$. On this range we estimate $L'$ by the levels bracketed by both curves, excluding levels at or below either curve's first
observed recall and those above either curve's maximum. \emph{Token-matched} multipliers use the final BF16
checkpoint as reference. \emph{Loss-matched} multipliers use the BF16 recall curve interpolated along its trajectory
at the low-precision run's final validation loss, interpolating linearly in validation loss between BF16 evaluations
(every 100 of about 2,000 steps on synthetic data and every 1,000 of 30,517 on natural text). No comparison is made
outside the observed BF16 loss range. The low-precision run and its reference see every fact on the same schedule,
with its exposures spread uniformly over training. The multipliers are horizontal shifts across exposure bins and do
not directly measure the wall-clock time that native FP4 hardware would save. We report the geometric mean over seeds
with a 95\% $t$-interval on the $\log_2$ scale, given in brackets. Since the probed facts and the data order are fixed,
these intervals reflect the randomness of initialization and training, and resampling the probed persons as well
changes them little (Appendix~\ref{app:robust}). Comparisons between configurations are paired by seed. Our primary
comparisons are \nvfp{} against BF16 in the two regimes. They and their main controls use six seeds, and the
split-moment and learning-rate tests with noise at twice the \nvfp{} error use eight because their effects are
smaller. Secondary variants (other noise levels, splits at the \nvfp{} error, single-operand quantization, training
settings, the natural-text recipe and analyses of saved checkpoints) use at least three. Table~\ref{tab:robust} and
Appendix~\ref{app:prereg} give the multiplier, interval and seed count of every configuration. The text gives
intervals only for the main results, and those outside the primary comparisons are not adjusted for multiple
comparisons.

\section{FP4 Needs More Exposures per Fact}\label{sec:main}
\begin{figure}[!t]
\centering
\includegraphics[width=\columnwidth]{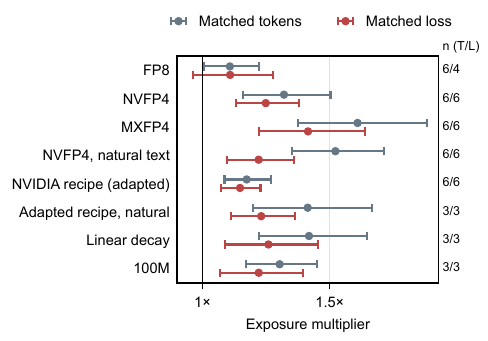}
\caption{Exposure multipliers against BF16 at matched tokens (gray) and matched validation loss (red), with 95\%
intervals and seed counts at matched tokens/loss (T/L). Values above 1 mean more exposures are needed. 40M synthetic
data unless labeled otherwise.
Further settings are in Table~\ref{tab:robust}.}
\label{fig:mult}
\end{figure}

Figure~\ref{fig:overview}(b, d) shows the final recall curves, and Figure~\ref{fig:mult} the resulting multipliers. On
synthetic data (40M) \nvfp{} shifts the curve right by 1.32$\times$ [1.16, 1.50] at matched tokens and 1.25$\times$
[1.13, 1.38] at matched validation loss, with every seed above 1 (sign test $p=0.03$), while the final validation loss
differs by only 0.1--0.9\%. On natural text the validation loss gap is larger (+1.9 to +2.1\%) and so is the
token-matched shift, 1.52$\times$ [1.35, 1.71]. At matched loss it is 1.22$\times$ [1.10, 1.36], again with all six seeds above 1 in both comparisons. On
natural text, where facts seen eight times or fewer are rarely learned, the recall levels are crossed between about
10 and 100 exposures, so the multiplier describes the learnable part of the tail. The natural-text loss-matched
BF16 point lies inside the
learning-rate decay
(step 28{,}700--28{,}900 of 30{,}517), where BF16 has seen 5--6\% fewer tokens and has not finished the decay. The
loss-matched control therefore compares different training stages, rather than removing all differences except precision.
Both lower BF16 recall at that point, biasing the loss-matched multiplier towards 1.

\paragraph{How the delay depends on the recall level (P1).}
Measured at single recall levels of 30\%, 50\% and 70\% (matched tokens, Appendix~\ref{app:robust}), the \nvfp{}
multiplier is 1.44/1.52/1.48$\times$ on natural text and 1.38/1.34/1.26$\times$ on synthetic data. On synthetic data
the multiplier is larger at 30\% than at 70\% in 18 of 19 settings, typically by a quarter on the $\log_2$ scale
(the ratio of the $\log_2$ shifts at 70\% and 30\% is 0.72 for \nvfp{}, 0.69 for backward-only quantization and 0.91 for i.i.d.\ noise),
whereas an exact shift of the BF16 curves, read on our exposure grid, gives 0.93. The delay is therefore approximately
a constant factor. On synthetic data it is largest at low recall levels, which the curves reach at the fewest
exposures, and this is the direction in which the checkpoint diagnostic predicts slightly larger $m_f$ for rarer
facts (Section~\ref{sec:mech}). The multiplier averaged over levels summarizes this shift.

\begin{figure*}[!t]
\centering
\includegraphics[width=\textwidth]{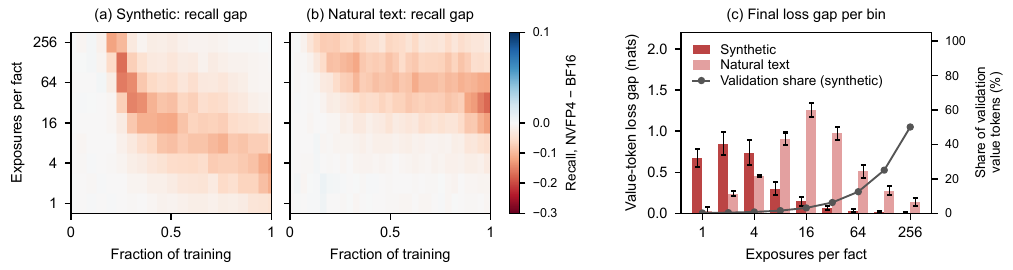}
\caption{Where the deficit sits (40M, mean over six seeds per regime). (a, b) Recall of \nvfp{} minus BF16 per
exposure bin during training (red: \nvfp{} behind). (c) Final value-token loss gap per bin ($\pm$1 standard error
over seeds) and the share of synthetic validation value tokens in each bin (right axis).}
\label{fig:acq}
\end{figure*}

\paragraph{The deficit follows the learning front.}
Figure~\ref{fig:acq}(a, b) shows when and where \nvfp{} falls behind. On synthetic data the deficit appears first
for facts seen 256 times, a fifth of the way into training, and moves to rarer facts as training proceeds. It is
largest at 16--32 exposures by mid-training and at 2--4 exposures at the end (0.11--0.12 recall), while facts seen
32 times or more end within 0.01. On natural text it sits at 64--128 exposures for most of training and at 32 at
the end, growing from 0.10 to 0.17 during the learning-rate decay. A delay predicts this pattern, because the recall
difference is largest where the learning curve is steepest and that point moves to rarer facts over training.

\paragraph{Loss sees the delay weighted by frequency.}
Stratifying the loss of the value tokens by exposure count shows the weighting of Proposition~\ref{prop:loss}
directly (Figure~\ref{fig:acq}(c)). On synthetic data \nvfp{} raises it by 0.67--0.85 nats for facts seen 1--4 times and by at most 0.07 nats
for facts seen 32 times or more. Averaged with equal weight per fact the gap is 0.32 nats. Averaged with the training
frequency, as in the validation stream, it is 0.04 nats, because half of the value tokens belong to facts
seen 256 times, which both models recall. Over all validation tokens the loss rises by 0.5\%. On natural text, where
facts need more exposures, frequency weighting shrinks the gap less (0.53 to 0.32 nats), and the validation loss,
measured on web text, contains none of the inserted biographies. The two regimes thus test two limitations of
validation loss. The synthetic stream contains the facts in proportion to their frequency and isolates the weighting
of Proposition~\ref{prop:loss}. The natural-text validation set, as in practice, lacks the rare facts a model should
acquire.

\paragraph{Matched tokens versus matched loss.}
For plain \nvfp{}, token-matched estimates exceed 1 under every tested readout, with intervals that exclude 1. Reading the \nvfp{} weights with BF16 forward GEMMs, which removes quantization error at evaluation time,
gives 1.26$\times$ on synthetic data and 1.30$\times$ on natural text (Appendix~\ref{app:robust}). Evaluation-time
quantization thus accounts for part of the default-readout gap, 15\% and 37\% of its $\log_2$ size, but not for all
of it. The loss-matched cost is more fragile. BF16-forward evaluation gives 1.21$\times$ (synthetic) and 1.11$\times$ [0.97, 1.26] (natural
text), and scoring the probability of the correct value gives 1.19$\times$ and 1.09$\times$. Part of the loss-matched gap, especially on natural text, is thus quantization error at evaluation
time rather than missing knowledge. We treat the token-matched multiplier as the primary measure
of acquisition and the loss-matched one as a control for the loss gap.

\section{Mechanism: Gradient Noise Through Adam}\label{sec:mech}
\begin{figure*}[!t]
\centering
\includegraphics[width=\textwidth]{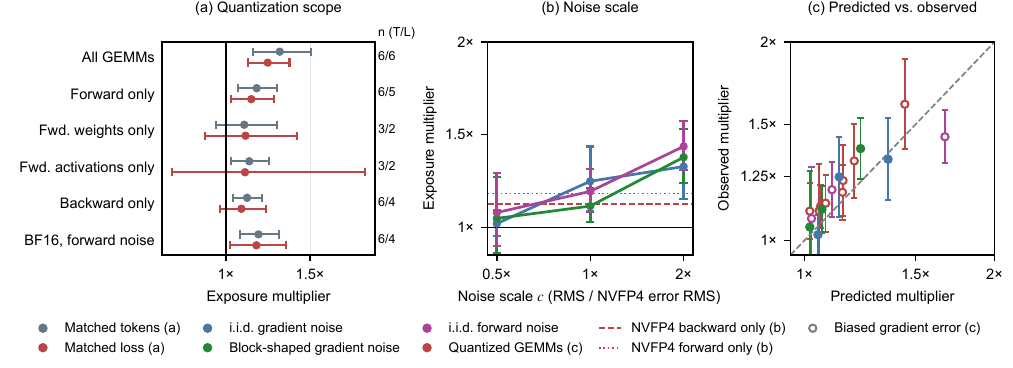}
\caption{Where the cost arises (40M synthetic, 95\% intervals). (a) Partial quantization and forward noise at
matched tokens (gray) and matched loss (red), with seed counts (T/L). (b) Zero-mean noise at 0.5, 1 and 2 times the
\nvfp{} error on backward GEMM outputs, i.i.d.\ (blue) or block-shaped (green), or on forward GEMM outputs (purple),
at matched tokens. Dashed and dotted: \nvfp{} backward-only and forward-only quantization. (c) Multiplier predicted
by Eq.~\eqref{eq:mf} against the observed token-matched multiplier for the 17 configurations of
Table~\ref{tab:theory}. Open markers: bias ratio above 0.1 (Appendix~\ref{app:precond}). Dashed: equality.}
\label{fig:mech}
\end{figure*}

\paragraph{Gradient error costs exposures.}
At matched tokens, quantizing only the forward GEMMs costs 1.18$\times$ and only the backward GEMMs 1.13$\times$
[1.04, 1.21], with every seed above 1 in both, and the two costs approximately multiply to the full effect ($1.18\times1.13=1.33$ against 1.32$\times$). Backward-only quantization runs the forward pass in BF16 and
changes validation loss by just $+0.1\%$ on average, so its cost is nearly invisible in the loss. Its gradient error
is approximately zero-mean (stochastic rounding, Appendix~\ref{app:precond}), and exactly zero-mean noise reproduces
its cost. We add Gaussian noise to
the outputs of both backward GEMMs of BF16 training, with the per-tensor RMS of the \nvfp{} error of each GEMM, and
scale it by the maximum magnitude of each 16-element block as in block-scaled rounding. This block-shaped noise costs
1.11$\times$, and 0.99$\times$ [0.94, 1.04] relative to backward-only quantization (paired by seed). The same RMS
spread uniformly over each tensor costs more, 1.25$\times$, or 1.12$\times$ [1.01, 1.24] relative to block-shaped
noise. This is the ordering of P3, since uniform noise places relatively more
error where the gradient is small. The cost grows with the noise scale (Figure~\ref{fig:mech}(b)). Block-shaped
noise at 0.5, 1 and 2 times the \nvfp{} error costs 1.05$\times$, 1.11$\times$ and 1.38$\times$, and i.i.d.\ noise
costs 1.02$\times$, 1.25$\times$ and 1.33$\times$ (Table~\ref{tab:robust}). Zero-mean gradient error is therefore
sufficient to cause an exposure cost. Running the backward GEMMs of \nvfp{} in FP8 lowers its cost from 1.32$\times$
to 1.23$\times$, close to forward-only quantization. Quantizing only the forward weights or activations costs
1.11$\times$ or 1.14$\times$.

\begin{figure*}[!t]
\centering
\includegraphics[width=\textwidth]{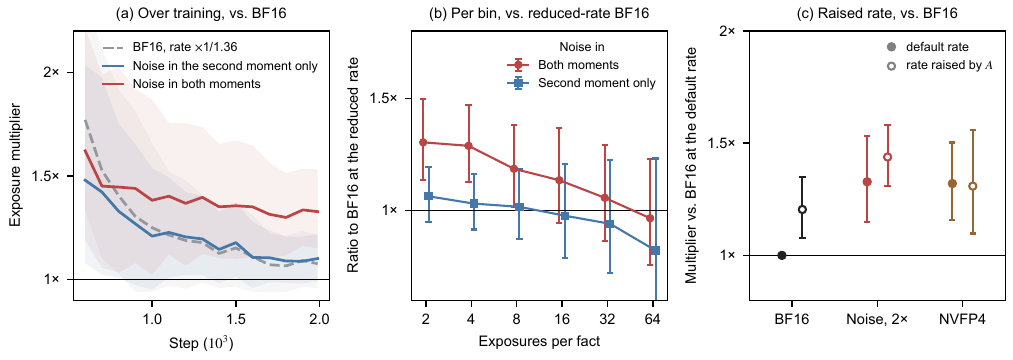}
\caption{Splitting gradient noise between Adam's moments (40M synthetic, i.i.d.\ noise at twice the \nvfp{} error,
8 seeds, 95\% intervals). (a) Token-matched multiplier against BF16 at $10^{-3}$, recomputed at every evaluation, for
BF16 at $10^{-3}/A$ (gray), noise in the second moment only (blue) and noise in both moments (red). (b) Horizontal
gap per exposure bin against BF16 at $10^{-3}/A$, averaged over evaluations (Appendix~\ref{app:prereg}). (c)
Multipliers against BF16 at $10^{-3}$ for training at $10^{-3}$ (filled) and at the rate raised by $A$ (open).}
\label{fig:split}
\end{figure*}

\paragraph{Adam turns gradient noise into two costs.}
Proposition~\ref{prop:adam} distinguishes two effects of zero-mean gradient error under Adam. A larger normalizer
shrinks the expected step like a lower learning rate, and noise in the update lowers its signal-to-noise ratio. We
separate them by splitting the i.i.d.\ noise configuration at the optimizer, at twice the \nvfp{} error where both
effects are large. At every step a second backward pass computes the BF16 gradient of the same batch, scaled by the
global-norm clipping factor of the noisy gradient, and Adam's two moments receive different gradients
(Proposition~\ref{prop:adam}(iii)). The shared factor barely couples them. In seed-0 reruns that log the gradient
norm, clipping acts on 3--4\% of steps for each configuration, mostly early in training, and in the second-moment-only
variant on the same steps as it would for the BF16 gradient (Appendix~\ref{app:prereg}).

With the noisy gradient in the second moment only, this variant acts as a learning-rate cut. Before training it, we
fixed the factor $A=1.356$ that Eq.~\eqref{eq:mf} predicts for this noise and trained BF16 at $10^{-3}/A$, holding
the product of learning rate and weight decay fixed. The second-moment-only variant closely matches this reduced-rate
BF16. At the end of training it costs 1.02$\times$ [0.93, 1.13] relative to it, an interval that excludes
differences larger than 13\%. The paired ratio stays within 0.93--1.03 at every evaluation from step 700 on
(Figure~\ref{fig:split}(a)), and no exposure bin shows a detectable difference (Figure~\ref{fig:split}(b)). The cost of such a
cut fades over training. Against BF16 at $10^{-3}$, the reduced rate costs 1.08$\times$ [0.96, 1.21] at the end of
training, and within each exposure bin its gap shrinks by 0.36 [0.19, 0.54] $\log_2$ exposures per thousand steps.
This match to a uniform cut did not hold above the default rate. At $2\times10^{-3}$, the second-moment-only variant
costs 1.66$\times$ [1.33, 2.07] more than BF16 at $2\times10^{-3}/A$, against a pre-specified prediction of no
difference (Appendix~\ref{app:prereg}). The cut is per coordinate and spares coordinates without injected error, such
as the output head. Where BF16's cost rises steeply with the rate, as it does above $10^{-3}$, those coordinates may
set the cost.

Noise in the update carries the rest of the cost, and over our runs this part does not fade, including during the
learning-rate decay to zero in the last 20\% of training. Against the reduced-rate BF16, noise in both moments costs
1.22$\times$ at step 1{,}600, as the decay begins, and 1.23$\times$ at the end. At equal expected step, noise in both
moments costs 1.21$\times$ [1.05, 1.39] more than noise in the second moment only (paired, $p=0.017$, 7 of 8 seeds),
two thirds of its $\log$-cost of 1.33$\times$ (seed bootstrap 38--91\%). These shares describe zero-mean
noise at twice the \nvfp{} error. Full \nvfp{} training adds biased gradient error (Table~\ref{tab:theory}), so they
may not transfer to it. Against the reduced-rate BF16, the gap of noise in both moments shows no trend within an
exposure bin (+0.06 [$-$0.09, 0.21] $\log_2$ per thousand steps) but grows as facts become rarer. At the same step,
every halving of a fact's exposures widens it by 0.08 [0.04, 0.12] $\log_2$ exposures, from 0.97$\times$ for facts
seen 64 times to 1.30$\times$ for facts seen twice (Figure~\ref{fig:split}(b)). Because rarer
facts are also less learned at any step, this profile does not separate exposure count from recall level. With the
noise in the first moment only, the normalizer no longer grows with the noise, but the update does. At the \nvfp{}
error this costs 1.06$\times$ (Table~\ref{tab:robust}), and at twice that error 6 of 8 runs diverged (validation loss
rose by more than one nat above its running minimum). Correcting Adam's normalizer for the error thus removes the
step-size cost but enlarges the update noise.

If the cost were a uniformly shrunken step, a larger learning rate would remove it. Raising the rate by $A$, with the
product of rate and decay fixed, restores the average expected step of part (i) under the local model but leaves the
signal-to-noise ratio of part (ii) unchanged. We fixed the predictions of this test before the runs (Appendix~\ref{app:prereg}). The cost does not fall
(Figure~\ref{fig:split}(c)). Relative to its multiplier at $10^{-3}$, the noise at the raised rate costs 1.08$\times$
[0.95, 1.24], against 0.75 if the cost were the shrunken step. \nvfp{} at a rate raised by its own factor $A=1.199$
costs 0.99$\times$ [0.87, 1.13], against 0.76. At the raised rate the noise costs 1.19$\times$ against BF16 at the
same rate, close to the update-noise cost at the default rate. The test rules out a
cost that consists only of a uniformly shrunken step. A larger rate also changes the trajectory in other ways, which this test
does not separate.

\paragraph{A fit-free checkpoint diagnostic.}
Eq.~\eqref{eq:mf} needs only quantities measured on a BF16 checkpoint. On each of three BF16 checkpoints (40M) we
estimate $v_i$ from 32 training batches, $s_i^2$ for every configuration as the mean squared difference
between its gradient and the BF16 gradient of the same batch, and $g_f$ for 96 facts seen 4--16 times
(Appendix~\ref{app:precond}). Nothing is fitted to observed multipliers. Across 17 configurations (formats, partial
quantization, and i.i.d., block-shaped and forward noise at three scales each), the prediction ranks the observed
costs with Spearman correlation 0.96 (bootstrap interval over configurations [0.84, 0.99]). The correlation is 0.95
without the 0.5$\times$ and 2$\times$ scales, whose order Eq.~\eqref{eq:mf} fixes by construction, and 0.89 among the
seven configurations with zero-mean gradient error. The prediction lies inside the observed 95\% interval for 15 of
the 17 (Figure~\ref{fig:mech}(c), Table~\ref{tab:theory}). It ranks the costs but does not calibrate them. For
zero-mean error it matches i.i.d.\ noise at twice the \nvfp{} error (1.36 predicted, 1.33$\times$ observed) but
underestimates smaller errors (1.13 against 1.25$\times$ for i.i.d.\ noise at the \nvfp{} error, 1.06 against
1.13$\times$ for backward-only quantization), and its value depends on how the normalizer is estimated
(Appendix~\ref{app:precond}). The split-moment variants deviate in the directions Proposition~\ref{prop:adam}
implies. Eq.~\eqref{eq:mf} sees only the normalizer, so it assigns the second-moment-only variant the full prediction
of its noise (1.36 against 1.10$\times$ observed) and the first-moment-only variant none (1.00 against 1.06$\times$).
For \nvfp{} the predicted $m_f$ vary little across facts (10th--90th percentile 1.16--1.26$\times$) and are slightly
larger for rarer facts (1.22/1.19/1.19$\times$ at 4/8/16 exposures). Configurations that quantize or perturb the
forward pass have biased gradient error, outside Assumption~\ref{ass:acc}(a), and for them the diagnostic tracks the
cost without explaining it.
Checkpoint seeds agree closely (standard deviation of $\log_2$ predictions at most 0.06). We have tested the
diagnostic only at one model size and within this training regime.

\paragraph{Zero-mean forward noise.}
Zero-mean Gaussian noise on the outputs of the BF16 forward GEMMs, with the per-tensor RMS of the \nvfp{} forward
error and a BF16 backward pass, costs 1.19$\times$ at matched tokens (every seed above 1, evaluated without the
noise), about as much as forward-only quantization. A cost thus does not require systematic
error at the perturbed output. Forward noise does, however, bias the gradient error
(Appendix~\ref{app:precond}), so this control cannot isolate the zero-mean mechanism.

\paragraph{Lenient readouts recover little of the deficit.}
A model could hold a fact yet fail to generate it. At the position before the answer, where factual associations are
retrieved~\citep{geva2023dissecting,meng2022rome}, we score the candidate values of the attribute at the output and,
through the final norm and unembedding, after each block (the logit
lens~\citep{nostalgebraist2020logitlens,belrose2023tunedlens}). A top-five readout at the output tests whether correct
values fall just below the top prediction. On three seeds rerun with saved checkpoints, it reduces the multiplier
only modestly, from 1.36$\times$ to 1.30$\times$ for \nvfp{}, from 1.21$\times$ to 1.17$\times$ for the adapted
recipe and from 1.18$\times$ to 1.14$\times$ for i.i.d.\ noise (Figure~\ref{fig:probe} in Appendix~\ref{app:controls}).
It preserves 71--92\% of the log-exposure shift in all nine pairs of seed and configuration. Correct-value probability
gives similar token-matched multipliers (1.32$\times$ synthetic, 1.46$\times$ natural text, Appendix~\ref{app:robust}).
Other probes might still find the knowledge, and Appendix~\ref{app:controls} gives intermediate-layer results and
probe limitations.

\section{Mitigation: Loss Recovers Faster Than Recall}\label{sec:recipe}
\begin{figure}[!t]
\centering
\includegraphics[width=\columnwidth]{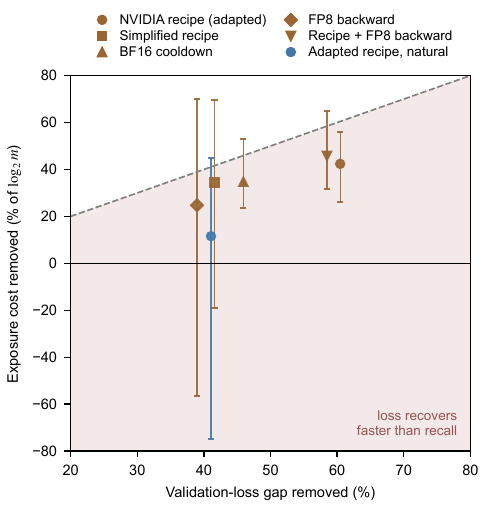}
\caption{Share of the validation-loss gap (horizontal) and of the $\log_2$ exposure cost at matched tokens
(vertical) of plain \nvfp{} that each remedy removes, on shared seeds (six synthetic, three natural). Bars: 95\%
Fieller intervals for the cost share (Appendix~\ref{app:controls}). Below the diagonal, loss recovers faster than
factual acquisition. Loss-matched values are in Table~\ref{tab:recipe}.}
\label{fig:recovery}
\end{figure}
\paragraph{Tested interventions.}
We test the interventions a practitioner would reach for first, each against plain \nvfp{} on the same seeds:
backward GEMMs in FP8; BF16 for the learning-rate decay (the last 20\% of steps, `BF16 cooldown'); a simplified
recipe (BF16 in the last 2 of 8 blocks, RHT also on the data gradient); the published \nvfp{} recipe adapted to our models;
and that recipe with FP8 backward GEMMs. All of them lower the estimated exposure cost, and none removes it. Every
remedy's token-matched multiplier stays above 1 with an interval that excludes 1.
We adapt the published recipe~\citep{nvidia2025nvfp4} as follows: 16$\times$16 two-dimensional scaling for weights,
so the forward
and data-gradient GEMMs see the same quantized weight; stochastic rounding only for gradients (nearest rounding for
weights and activations); RHT only on the weight-gradient inputs; and high-precision final layers (we
keep 2 of 8 blocks in BF16, more than its $\sim$15\%). Figure~\ref{fig:recovery} shows that this adaptation removes
60\% of the validation loss gap but 42\% of the $\log_2$ exposure cost at matched tokens (37\% at matched loss,
where the remaining 1.15$\times$ is above 1 in all six seeds, Table~\ref{tab:recipe}). Training the whole
learning-rate decay in BF16, which the recipe suggests for closing the loss gap, running the backward GEMMs in FP8,
and a simplified recipe fall in the same region. Every remedy we tried lies below the diagonal, where loss recovers
faster than factual acquisition. When seeds are resampled jointly for both shares, the difference between them
excludes zero for the adapted recipe (18 percentage points [2, 29]) and for the BF16 cooldown (11 [3, 29]). For the
other remedies its interval is wide.
Running the recipe's backward GEMMs in FP8 changes little (1.16$\times$ against 1.17$\times$ at matched tokens).
The BF16 cooldown does not repair facts seen before it. Among facts seen at most 8 times, those whose exposures all
precede it keep the full \nvfp{} recall deficit (cooldown minus plain \nvfp{}: $-0.003$ [$-0.017$, 0.011], about
2{,}400 facts per seed). Those seen only during the cooldown keep 54\% [36, 70] of the deficit (288 facts per seed),
and facts seen on both sides of the switch are excluded (Appendix~\ref{app:controls}). An \nvfp{}-trained model thus
continues to learn new facts less efficiently after switching to BF16. This suggests a training-history effect, such as a later formation of the shared circuits that
support recall~\citep{zucchet2025facts}, in addition to the noise of the current update.

\paragraph{Natural text.}
In the point estimates, the discrepancy is larger on natural text (blue point). Paired with the same BF16 runs, the
adapted recipe shrinks the validation-loss gap from 1.96\% to 1.15\% but changes the exposure cost little, at 1.41$\times$ at
matched tokens against 1.48$\times$ for plain \nvfp{} on the same seeds, and 1.23$\times$ at matched validation loss
against 1.16$\times$, above 1 in all three seeds. The larger loss-matched estimate
does not imply worse acquisition. Because the recipe ends closer to BF16 in loss, its loss-matched BF16 checkpoint is
a later one, so it is compared against a better-trained reference. At matched tokens it removes about an eighth of
the excess cost ($\log_2$ scale), 12\% [$-$75, 45] over three seeds (Figure~\ref{fig:recovery}).

\paragraph{Directions suggested by the model.}
Proposition~\ref{prop:adam} places the cost in the gradient error relative to the gradient noise,
$\rho_i=s_i^2/v_i$, and Section~\ref{sec:mech} finds most of it in the update's signal-to-noise ratio, which a
change of step size does not restore. Remedies should therefore lower $\rho_i$ where it is largest rather than
rescale the step. Two routes follow from our results. First, the error can be shaped to follow the gradient's scale.
At equal RMS, block-shaped gradient noise costs 1.11$\times$ against 1.25$\times$ for i.i.d.\ noise. Second,
precision can be spent on the backward pass, whose cost loss barely registers. FP8 backward GEMMs lower the cost of
\nvfp{} from 1.32$\times$ to 1.23$\times$. Correcting Adam's normalizer for the error, for instance from the known
variance of stochastic rounding, does not help. An oracle version removes the step-size part but enlarges the update
noise, and at twice the \nvfp{} error it destabilized 6 of 8 runs. Finally, since the cost acts approximately as a
constant delay, sampling knowledge-dense data about $m$ times more often would restore the number of exposures,
although its effect on other capabilities is untested.

\section{Robustness Across Training Settings}\label{sec:scale}
\begin{figure}[!t]
\centering
\includegraphics[width=\columnwidth]{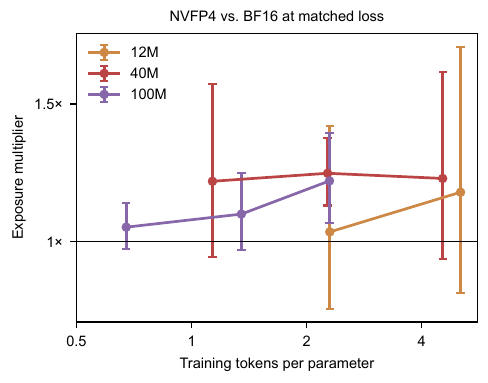}
\caption{Loss-matched exposure multiplier of \nvfp{} against BF16 vs.\ training tokens per parameter for the 12M,
40M and 100M architectures (12.9M, 28.7M and 96.6M parameters with the synthetic vocabulary, all counted) on
synthetic data, with 95\% intervals over 3--6 seeds per point.}
\label{fig:scale}
\end{figure}
\paragraph{Precision and training budget.}
The cost increases as precision decreases. Paired by seed, the matched-token multiplier of \nvfp{} is 1.19$\times$
[1.02, 1.40] that of FP8, and that of MXFP4 is 1.22$\times$ [1.01, 1.47] that of \nvfp{}. FP8 training shows a small
cost near the measurement resolution, 1.11$\times$ at matched tokens and at matched loss, with intervals that touch or
include 1 (Table~\ref{tab:robust}). Two of its seeds end below the BF16 validation loss and have no loss-matched
point. MXFP4 costs more than \nvfp{} (1.61$\times$ vs.\ 1.32$\times$ at matched tokens, 1.41$\times$ vs.\ 1.25$\times$
at matched loss). At 100M the loss-matched cost grows with training tokens per parameter, from 1.05$\times$ at 0.67 to
1.10$\times$ at 1.35 and 1.22$\times$ at 2.3, where it matches 40M. Because the synthetic stream grows
with the number of persons ($N$ from 9{,}100 to 68{,}000), tokens per parameter and facts per parameter change
together in Figure~\ref{fig:scale}, so these trends reflect training budget and memorization load jointly. At 40M we
see no trend between 1.1
and 4.5 tokens per parameter. At 12M seed variance exceeds the effect. Production models train far beyond these
ratios, and our natural-text runs (about 24 tokens per parameter) are the closest point. The cost also persists under
a different schedule. With linear decay over the whole run it is 1.26$\times$ at matched loss and 1.42$\times$ at
matched tokens, above 1 in every seed.

\paragraph{Learning-rate sensitivity.}
We also train both formats at peak rates $5\times10^{-4}$ and $2\times10^{-3}$ (40M, three seeds). Selecting each
format's rate by mean final recall over facts seen 1--8 times selects $10^{-3}$ for both. On these selection seeds
the token-matched multiplier is 1.37$\times$. On three further seeds, which played no part in the selection, it
is 1.27$\times$ [1.02, 1.58] at matched tokens and 1.20$\times$ [1.01, 1.44] at matched loss, so the cost survives
rate selection, although a three-point sweep cannot exclude an untested rate. \nvfp{} at $5\times10^{-4}$ performs
comparably but has lower mean recall. Its 1.47$\times$ comparison to BF16 at $10^{-3}$ uses an alternative rate for
\nvfp{} and does not compare best-recall settings (Appendix~\ref{app:controls}). This sweep keeps the weight decay
coefficient fixed, so the decay per step grows with the rate. Proposition~\ref{prop:adam} motivates a sharper test
that holds the product of rate and decay fixed. Raising the rate by $A$ then restores the average expected step of
part (i) but not the signal-to-noise ratio of part (ii). Raising the rate of \nvfp{} by its $A=1.199$ in this way leaves its
cost unchanged, at 0.99$\times$ of its multiplier at $10^{-3}$ (pre-specified, Section~\ref{sec:mech}), and at the
raised rate it costs 1.31$\times$ against BF16 at $10^{-3}$.
Neither test finds a learning rate at which the cost disappears.

\section{Related Work}
\paragraph{Low-precision training.}
Reduced-precision training rests on two ideas that our configurations inherit, a high-precision master copy of the
weights~\citep{micikevicius2018mixed} and stochastic rounding, which makes quantization
unbiased~\citep{gupta2015limited}.
FP8 training is now standard at scale~\citep{micikevicius2022fp8,peng2023fp8lm,fishman2025fp8,deepseek2024v3}. Earlier
4-bit training work used integer formats with Hadamard transforms against outliers~\citep{xi2023int4} and unbiased,
logarithmically scaled quantization of the neural gradients~\citep{chmiel2021fp4}. Current
FP4 recipes combine fine-grained block scaling~\citep{rouhani2023mx}, stochastic rounding of gradients, random
Hadamard rotations and high-precision layers~\citep{nvidia2025nvfp4,wang2025fp4,tseng2025mxfp4,castro2025quartet,chmiel2025fp4}.
These methods are typically compared by loss gap and benchmarks~\citep{agrusa2026nvfp4,cao2026averis}, and precision-aware scaling laws
model low-precision training as a loss in effective parameter count~\citep{kumar2024precision}. All of these are
loss-based or aggregate evaluations. We instead measure the exposures required for factual recall and show that
this criterion responds differently to precision and recipe changes.
\paragraph{Quantization and knowledge.}
Post-training compression can preserve perplexity while degrading knowledge-intensive tasks~\citep{jaiswal2023compressing},
and its damage grows with the amount of pretraining~\citep{kumar2024precision,ouyang2024lowbit}. Rotation-based and
second-order methods reduce it~\citep{frantar2023gptq,ashkboos2024quarot}. In controlled settings, language models
store about 2 bits of knowledge per parameter after enough exposures, and int4 post-training quantization reduces
this to about 0.7 bits~\citep{allenzhu2024capacity}. Closest to us, ternary-weight training stores far fewer facts
than FP16 while perplexity rises only moderately, through a learning-rate-dependent collapse localized in the output
head~\citep{paudel2026linger}. We study emulated versions of the FP4 formats used in production, with both passes quantized and a
high-precision output head. Our question is exposure efficiency rather than retained capacity. We observe a
rightward shift of the recall curve without the collapse reported for ternary training, and we locate part of it in
the gradients, since quantizing only the backward pass, with a BF16 forward pass, already costs exposures. As in that
work, the deficit arises during training, unlike damage introduced by post-training compression.
\paragraph{Acquiring facts during pretraining.}
Language models answer factual queries from their parameters~\citep{petroni2019lama}, but their accuracy depends
strongly on how often a fact appears in pretraining~\citep{kandpal2023longtail,mallen2023popqa}. Controlled studies
show that facts are acquired through repeated exposures and partly forgotten in between~\citep{chang2024acquire}, that
memorization depends on model size and repetition~\citep{tirumala2022memorization,carlini2023memorization}, and that
synthetic biographies with diverse phrasings make knowledge extractable~\citep{allenzhu2023physics31}. We use this
setting as a measuring instrument for numerical precision. Our noise experiments relate it to work that measures how
gradient noise limits the progress of each optimization step~\citep{mccandlish2018largebatch} and to analyses of
Adam in which the per-coordinate step shrinks as the relative variance of the gradient grows~\citep{balles2018dissecting,malladi2022sde}.
\citet{chmiel2025fp4} relate the effectiveness of FP4 training to the ratio of gradient norm to quantization noise.
Eq.~\eqref{eq:mf} is a per-coordinate counterpart of that ratio, weighted by the gradient of a fact, and we relate it
to the exposures a fact needs. Gradient noise added for differential privacy likewise lowers accuracy most on
underrepresented classes~\citep{bagdasaryan2019dp}. We show how Adam turns such noise into a cost per exposure.

\section{Discussion}\label{sec:discussion}
\paragraph{What the model explains.}
The model is an optimizer-level account supported by interventions and does not provide a closed-form theory of
factual acquisition. It separates two costs of zero-mean gradient error under Adam and predicts how each responds to
the split-moment intervention. The step-size cost matched a learning-rate cut fixed in advance at the default rate,
but not above it, and a larger rate that restores the average expected step did not reduce the cost, as the signal-to-noise
part implies. Its quantitative part, Eq.~\eqref{eq:mf}, prices only the expected step. It ranks configurations but
underestimates small errors, and the signal-to-noise cost, which carries two thirds of the measured cost at twice the
\nvfp{} error, has no closed-form multiplier in our account. The linearized model explains why one cost fades and
the other persists over our runs, but its link from weight spread to recall is heuristic. The account is local and
omits forgetting between exposures~\citep{chang2024acquire}, curvature and interactions between facts. A later
formation of the shared attention circuits that support recall~\citep{zucchet2025facts} would also shift every
exposure bin. The BF16 cooldown, after which facts seen only in BF16 keep about half of the deficit, is consistent
with such training-history effects, and our experiments do not separate them from slower progress on each fact. Finally, the
evidence that zero-mean gradient noise causes the cost concerns its backward-pass part. The correlation across 17
configurations does not establish that the same mechanism causes the cost of full FP4 training, whose gradient error
is biased.
\paragraph{Loss parity at scale.}
Reports of loss and downstream parity for FP4 pretraining at scale~\citep{nvidia2025nvfp4,chmiel2025fp4} can be
reconciled with these results, because token-averaged loss weights frequent content (Proposition~\ref{prop:loss}) and
aggregate benchmarks do not isolate rare facts. Higher throughput can offset the cost only insofar as the additional
tokens contain further occurrences of the relevant facts, from repeated data or new documents, and we do not measure
this trade-off. Whether the multiplier shrinks or grows
at production scale remains open (Section~\ref{sec:scale}).
\paragraph{Other optimizers.}
Only the step-size cost depends on coordinatewise normalization. SGD with momentum has no normalizer to inflate, so
under the same approximation zero-mean error leaves its expected step unchanged. The signal-to-noise cost does not
depend on normalization, since the error adds variance to any stochastic update, and under Adam it is the larger
part. Switching to SGD would therefore not necessarily avoid the cost. Matrix-based optimizers such as Muon~\citep{liu2025muon} require
a different analysis. Optimizer comparisons are a direct test for future work.
\paragraph{Implications for evaluating low-precision recipes.}
Our adapted \nvfp{} recipe recovers loss more effectively than factual acquisition, especially on natural text.
We therefore suggest reporting an exposure-controlled recall curve alongside the loss gap. This requires an
instrumented training corpus and a factual probe. In our natural-text runs, biographies occupy 7.2\% of tokens.
Equation~\eqref{eq:mf} offers an additional checkpoint diagnostic before retraining candidate recipes, although its
predictive value beyond these models remains untested.

\section{Limitations}
Our models have at most 96.6M parameters and see at most 1B tokens. Production-scale generalization remains open,
and the natural-text setting reaches about 24 tokens per parameter. FP4 GEMMs are emulated, with no native-hardware
validation or throughput measurement. Facts are synthetic biographies, on their own or mixed into one web corpus,
rather than naturally occurring long-tail knowledge. We test only AdamW, a three-rate sweep at 40M with three seeds,
and one alternative schedule. Forward-noise controls are i.i.d., and we do not test block-shaped forward noise.
Recall is probed with a template seen in training, and we do not test paraphrased questions. All runs share one data order, so the
seed intervals do not include its variability.
The model assumes zero-mean gradient error, which holds exactly for the noise controls and approximately for
backward-only \nvfp{} (Appendix~\ref{app:precond}), and fixed local statistics. Eq.~\eqref{eq:mf} uses final BF16
checkpoints at learning rate $10^{-3}$ rather than statistics along the training trajectory, its value depends on how
the normalizer is estimated, and it underestimates small errors. The split-moment intervention uses i.i.d.\ noise at
two scales, with clipping statistics from one rerun per configuration. Its second-moment-only variant matched a
uniform learning-rate cut at the default rate but not at $2\times10^{-3}$, and the profile of the signal-to-noise
cost over exposure counts does not separate exposure count from recall level. The compensation test raises the rate
by one factor per configuration.
Finally, loss-matched estimates depend on the readout. For plain \nvfp{} their intervals include 1 under
correct-value probability in both regimes and under BF16-forward evaluation on natural text
(Appendix~\ref{app:robust}), whereas token-matched intervals exclude 1 under every readout.

\section*{Impact Statement}
This paper studies how 4-bit training affects what language models learn from rarely seen data. A model that
acquires rare facts more slowly may be less reliable on knowledge about less frequent entities, topics and
languages, although we have not tested such effects. Measuring exposure costs may help
practitioners assess these risks alongside efficiency gains. Our experiments use fictitious biographies and a
public web corpus.

\bibliography{references}
\bibliographystyle{icml2026}

\newpage
\appendix
\onecolumn
\raggedbottom
\renewcommand{\topfraction}{0.9}
\renewcommand{\bottomfraction}{0.8}
\renewcommand{\textfraction}{0.07}
\renewcommand{\floatpagefraction}{0.85}
\section{Robustness of the Exposure Multiplier}\label{app:robust}
\begin{table}[H]
\centering
\caption{Default exposure multipliers averaged over recall levels 20--80\%, with geometric means, 95\% intervals in brackets and the number of seeds with a defined multiplier ($n$). Settings are grouped by the experimental factor changed.}
\label{tab:robust}
\footnotesize
\setlength{\tabcolsep}{2pt}
\renewcommand{\arraystretch}{1.0}
\begin{tabular*}{\textwidth}{@{\extracolsep{\fill}}lrrrr@{}}
\toprule
Setting & \multicolumn{2}{c}{Matched tokens} & \multicolumn{2}{c}{Matched loss} \\
\cmidrule(lr){2-3}\cmidrule(l){4-5}
 & $n$ & Multiplier [95\% CI] & $n$ & Multiplier [95\% CI] \\
\midrule
\addlinespace[4pt]
\multicolumn{5}{@{}l}{\itshape Formats} \\[-1pt]
FP8 & 6 & \mbox{1.11\,{\fontsize{7}{8}\selectfont\color{black!70}[1.00,\,1.22]}} & 4 & \mbox{1.11\,{\fontsize{7}{8}\selectfont\color{black!70}[0.96,\,1.28]}} \\
NVFP4 & 6 & \mbox{1.32\,{\fontsize{7}{8}\selectfont\color{black!70}[1.16,\,1.50]}} & 6 & \mbox{1.25\,{\fontsize{7}{8}\selectfont\color{black!70}[1.13,\,1.38]}} \\
MXFP4 & 6 & \mbox{1.61\,{\fontsize{7}{8}\selectfont\color{black!70}[1.38,\,1.88]}} & 6 & \mbox{1.41\,{\fontsize{7}{8}\selectfont\color{black!70}[1.22,\,1.64]}} \\
\addlinespace[4pt]
\multicolumn{5}{@{}l}{\itshape Recipes} \\[-1pt]
NVIDIA recipe (adapted) & 6 & \mbox{1.17\,{\fontsize{7}{8}\selectfont\color{black!70}[1.09,\,1.27]}} & 6 & \mbox{1.15\,{\fontsize{7}{8}\selectfont\color{black!70}[1.07,\,1.23]}} \\
Simplified recipe & 6 & \mbox{1.20\,{\fontsize{7}{8}\selectfont\color{black!70}[1.08,\,1.33]}} & 6 & \mbox{1.16\,{\fontsize{7}{8}\selectfont\color{black!70}[1.04,\,1.29]}} \\
BF16 cooldown & 6 & \mbox{1.20\,{\fontsize{7}{8}\selectfont\color{black!70}[1.08,\,1.33]}} & 5 & \mbox{1.18\,{\fontsize{7}{8}\selectfont\color{black!70}[1.09,\,1.29]}} \\
\addlinespace[4pt]
\multicolumn{5}{@{}l}{\itshape Natural-text mixture} \\[-1pt]
NVFP4 & 6 & \mbox{1.52\,{\fontsize{7}{8}\selectfont\color{black!70}[1.35,\,1.71]}} & 6 & \mbox{1.22\,{\fontsize{7}{8}\selectfont\color{black!70}[1.10,\,1.36]}} \\
NVIDIA recipe (adapted) & 3 & \mbox{1.41\,{\fontsize{7}{8}\selectfont\color{black!70}[1.20,\,1.66]}} & 3 & \mbox{1.23\,{\fontsize{7}{8}\selectfont\color{black!70}[1.11,\,1.36]}} \\
\addlinespace[4pt]
\multicolumn{5}{@{}l}{\itshape Backward precision} \\[-1pt]
NVFP4 + FP8 backward & 6 & \mbox{1.23\,{\fontsize{7}{8}\selectfont\color{black!70}[1.09,\,1.39]}} & 6 & \mbox{1.19\,{\fontsize{7}{8}\selectfont\color{black!70}[1.07,\,1.33]}} \\
Recipe + FP8 backward & 6 & \mbox{1.16\,{\fontsize{7}{8}\selectfont\color{black!70}[1.06,\,1.27]}} & 4 & \mbox{1.12\,{\fontsize{7}{8}\selectfont\color{black!70}[0.98,\,1.27]}} \\
\addlinespace[4pt]
\multicolumn{5}{@{}l}{\itshape Training settings} \\[-1pt]
Linear learning-rate decay & 3 & \mbox{1.42\,{\fontsize{7}{8}\selectfont\color{black!70}[1.22,\,1.65]}} & 3 & \mbox{1.26\,{\fontsize{7}{8}\selectfont\color{black!70}[1.09,\,1.46]}} \\
Peak rate $5\times10^{-4}$ & 3 & \mbox{0.91\,{\fontsize{7}{8}\selectfont\color{black!70}[0.38,\,2.16]}} & 3 & \mbox{0.85\,{\fontsize{7}{8}\selectfont\color{black!70}[0.36,\,1.97]}} \\
Peak rate $2\times10^{-3}$ & 3 & \mbox{1.49\,{\fontsize{7}{8}\selectfont\color{black!70}[0.70,\,3.18]}} & 3 & \mbox{1.38\,{\fontsize{7}{8}\selectfont\color{black!70}[0.74,\,2.55]}} \\
100M architecture & 3 & \mbox{1.30\,{\fontsize{7}{8}\selectfont\color{black!70}[1.17,\,1.45]}} & 3 & \mbox{1.22\,{\fontsize{7}{8}\selectfont\color{black!70}[1.07,\,1.40]}} \\
\addlinespace[4pt]
\multicolumn{5}{@{}l}{\itshape Quantization scope} \\[-1pt]
Forward only & 6 & \mbox{1.18\,{\fontsize{7}{8}\selectfont\color{black!70}[1.07,\,1.30]}} & 5 & \mbox{1.15\,{\fontsize{7}{8}\selectfont\color{black!70}[1.03,\,1.29]}} \\
Forward weights only & 3 & \mbox{1.11\,{\fontsize{7}{8}\selectfont\color{black!70}[0.94,\,1.31]}} & 2 & \mbox{1.12\,{\fontsize{7}{8}\selectfont\color{black!70}[0.88,\,1.42]}} \\
Forward activations only & 3 & \mbox{1.14\,{\fontsize{7}{8}\selectfont\color{black!70}[1.03,\,1.26]}} & 2 & \mbox{1.11\,{\fontsize{7}{8}\selectfont\color{black!70}[0.68,\,1.83]}} \\
Backward only & 6 & \mbox{1.13\,{\fontsize{7}{8}\selectfont\color{black!70}[1.04,\,1.21]}} & 4 & \mbox{1.09\,{\fontsize{7}{8}\selectfont\color{black!70}[0.96,\,1.24]}} \\
\addlinespace[4pt]
\multicolumn{5}{@{}l}{\itshape Noise controls} \\[-1pt]
Forward noise & 6 & \mbox{1.19\,{\fontsize{7}{8}\selectfont\color{black!70}[1.08,\,1.31]}} & 4 & \mbox{1.18\,{\fontsize{7}{8}\selectfont\color{black!70}[1.03,\,1.36]}} \\
i.i.d. gradient noise & 6 & \mbox{1.25\,{\fontsize{7}{8}\selectfont\color{black!70}[1.09,\,1.44]}} & 5 & \mbox{1.24\,{\fontsize{7}{8}\selectfont\color{black!70}[1.09,\,1.41]}} \\
Block-shaped gradient noise & 6 & \mbox{1.11\,{\fontsize{7}{8}\selectfont\color{black!70}[1.03,\,1.21]}} & 5 & \mbox{1.11\,{\fontsize{7}{8}\selectfont\color{black!70}[1.01,\,1.22]}} \\
\addlinespace[4pt]
\multicolumn{5}{@{}l}{\itshape Noise dose (multiple of the NVFP4 error)} \\[-1pt]
i.i.d. gradient noise, $0.5\times$ & 3 & \mbox{1.02\,{\fontsize{7}{8}\selectfont\color{black!70}[0.95,\,1.09]}} & 1 & 1.02 \\
i.i.d. gradient noise, $2\times$ & 8 & \mbox{1.33\,{\fontsize{7}{8}\selectfont\color{black!70}[1.15,\,1.53]}} & 8 & \mbox{1.25\,{\fontsize{7}{8}\selectfont\color{black!70}[1.11,\,1.41]}} \\
Block-shaped gradient noise, $0.5\times$ & 3 & \mbox{1.05\,{\fontsize{7}{8}\selectfont\color{black!70}[0.86,\,1.27]}} & 2 & \mbox{1.02\,{\fontsize{7}{8}\selectfont\color{black!70}[0.38,\,2.75]}} \\
Block-shaped gradient noise, $2\times$ & 3 & \mbox{1.38\,{\fontsize{7}{8}\selectfont\color{black!70}[1.24,\,1.53]}} & 3 & \mbox{1.31\,{\fontsize{7}{8}\selectfont\color{black!70}[1.26,\,1.36]}} \\
Forward noise, $0.5\times$ & 3 & \mbox{1.08\,{\fontsize{7}{8}\selectfont\color{black!70}[0.90,\,1.29]}} & 2 & \mbox{1.06\,{\fontsize{7}{8}\selectfont\color{black!70}[0.41,\,2.72]}} \\
Forward noise, $2\times$ & 3 & \mbox{1.44\,{\fontsize{7}{8}\selectfont\color{black!70}[1.31,\,1.58]}} & 3 & \mbox{1.32\,{\fontsize{7}{8}\selectfont\color{black!70}[1.17,\,1.51]}} \\
\addlinespace[4pt]
\multicolumn{5}{@{}l}{\itshape Split moments (i.i.d. gradient noise)} \\[-1pt]
Second moment only, $1\times$ & 3 & \mbox{1.06\,{\fontsize{7}{8}\selectfont\color{black!70}[0.96,\,1.17]}} & 1 & 1.08 \\
First moment only, $1\times$ & 3 & \mbox{1.06\,{\fontsize{7}{8}\selectfont\color{black!70}[1.00,\,1.11]}} & 2 & \mbox{1.05\,{\fontsize{7}{8}\selectfont\color{black!70}[0.79,\,1.39]}} \\
Second moment only, $2\times$ & 8 & \mbox{1.10\,{\fontsize{7}{8}\selectfont\color{black!70}[0.99,\,1.22]}} & 6 & \mbox{1.08\,{\fontsize{7}{8}\selectfont\color{black!70}[0.94,\,1.24]}} \\
\bottomrule
\end{tabular*}
\end{table}
\begin{table}[tbp]
\centering
\caption{Sensitivity to the recall level at which the exposure multiplier is computed. Geometric means with 95\% intervals. Seed counts are as in Table~\ref{tab:robust}, except where a cell ends with its own \emph{n}. --: unavailable or outside the BF16 loss range.}
\label{tab:robust-levels}
\footnotesize
\setlength{\tabcolsep}{2pt}
\renewcommand{\arraystretch}{1.0}
\begin{tabular*}{\textwidth}{@{\extracolsep{\fill}}lrrrrrr@{}}
\toprule
Setting & \multicolumn{3}{c}{Matched tokens} & \multicolumn{3}{c}{Matched loss} \\
\cmidrule(lr){2-4}\cmidrule(l){5-7}
 & 30\% & 50\% & 70\% & 30\% & 50\% & 70\% \\
\midrule
\addlinespace[4pt]
\multicolumn{7}{@{}l}{\itshape Formats} \\[-1pt]
FP8 & \mbox{1.13\,{\fontsize{7}{8}\selectfont\color{black!70}[1.02,\,1.26]}} & \mbox{1.11\,{\fontsize{7}{8}\selectfont\color{black!70}[1.00,\,1.22]}} & \mbox{1.11\,{\fontsize{7}{8}\selectfont\color{black!70}[0.98,\,1.25]}} & \mbox{1.14\,{\fontsize{7}{8}\selectfont\color{black!70}[0.96,\,1.35]}} & \mbox{1.11\,{\fontsize{7}{8}\selectfont\color{black!70}[0.96,\,1.28]}} & \mbox{1.10\,{\fontsize{7}{8}\selectfont\color{black!70}[0.94,\,1.29]}} \\
NVFP4 & \mbox{1.38\,{\fontsize{7}{8}\selectfont\color{black!70}[1.22,\,1.56]}} & \mbox{1.34\,{\fontsize{7}{8}\selectfont\color{black!70}[1.17,\,1.53]}} & \mbox{1.26\,{\fontsize{7}{8}\selectfont\color{black!70}[1.10,\,1.45]}} & \mbox{1.32\,{\fontsize{7}{8}\selectfont\color{black!70}[1.20,\,1.46]}} & \mbox{1.25\,{\fontsize{7}{8}\selectfont\color{black!70}[1.14,\,1.38]}} & \mbox{1.17\,{\fontsize{7}{8}\selectfont\color{black!70}[1.06,\,1.30]}} \\
MXFP4 & \mbox{1.70\,{\fontsize{7}{8}\selectfont\color{black!70}[1.44,\,1.99]}} & \mbox{1.62\,{\fontsize{7}{8}\selectfont\color{black!70}[1.41,\,1.87]}} & \mbox{1.46\,{\fontsize{7}{8}\selectfont\color{black!70}[1.24,\,1.72]}} & \mbox{1.53\,{\fontsize{7}{8}\selectfont\color{black!70}[1.31,\,1.77]}} & \mbox{1.40\,{\fontsize{7}{8}\selectfont\color{black!70}[1.20,\,1.62]}} & \mbox{1.25\,{\fontsize{7}{8}\selectfont\color{black!70}[1.09,\,1.43]}} \\
\addlinespace[4pt]
\multicolumn{7}{@{}l}{\itshape Recipes} \\[-1pt]
NVIDIA recipe (adapted) & \mbox{1.21\,{\fontsize{7}{8}\selectfont\color{black!70}[1.11,\,1.33]}} & \mbox{1.18\,{\fontsize{7}{8}\selectfont\color{black!70}[1.09,\,1.28]}} & \mbox{1.16\,{\fontsize{7}{8}\selectfont\color{black!70}[1.07,\,1.25]}} & \mbox{1.19\,{\fontsize{7}{8}\selectfont\color{black!70}[1.10,\,1.29]}} & \mbox{1.15\,{\fontsize{7}{8}\selectfont\color{black!70}[1.08,\,1.23]}} & \mbox{1.13\,{\fontsize{7}{8}\selectfont\color{black!70}[1.04,\,1.21]}} \\
Simplified recipe & \mbox{1.25\,{\fontsize{7}{8}\selectfont\color{black!70}[1.13,\,1.39]}} & \mbox{1.22\,{\fontsize{7}{8}\selectfont\color{black!70}[1.09,\,1.36]}} & \mbox{1.18\,{\fontsize{7}{8}\selectfont\color{black!70}[1.03,\,1.35]}} & \mbox{1.22\,{\fontsize{7}{8}\selectfont\color{black!70}[1.10,\,1.36]}} & \mbox{1.17\,{\fontsize{7}{8}\selectfont\color{black!70}[1.04,\,1.32]}} & \mbox{1.13\,{\fontsize{7}{8}\selectfont\color{black!70}[0.99,\,1.28]}} \\
BF16 cooldown & \mbox{1.24\,{\fontsize{7}{8}\selectfont\color{black!70}[1.11,\,1.38]}} & \mbox{1.21\,{\fontsize{7}{8}\selectfont\color{black!70}[1.08,\,1.36]}} & \mbox{1.18\,{\fontsize{7}{8}\selectfont\color{black!70}[1.06,\,1.32]}} & \mbox{1.23\,{\fontsize{7}{8}\selectfont\color{black!70}[1.13,\,1.35]}} & \mbox{1.20\,{\fontsize{7}{8}\selectfont\color{black!70}[1.11,\,1.29]}} & \mbox{1.15\,{\fontsize{7}{8}\selectfont\color{black!70}[1.05,\,1.26]}} \\
\addlinespace[4pt]
\multicolumn{7}{@{}l}{\itshape Natural-text mixture} \\[-1pt]
NVFP4 & \mbox{1.44\,{\fontsize{7}{8}\selectfont\color{black!70}[1.30,\,1.59]}} & \mbox{1.52\,{\fontsize{7}{8}\selectfont\color{black!70}[1.35,\,1.71]}} & \mbox{1.48\,{\fontsize{7}{8}\selectfont\color{black!70}[1.26,\,1.75]}} & \mbox{1.24\,{\fontsize{7}{8}\selectfont\color{black!70}[1.14,\,1.35]}} & \mbox{1.24\,{\fontsize{7}{8}\selectfont\color{black!70}[1.11,\,1.37]}} & \mbox{1.15\,{\fontsize{7}{8}\selectfont\color{black!70}[0.99,\,1.34]}} \\
NVIDIA recipe (adapted) & \mbox{1.37\,{\fontsize{7}{8}\selectfont\color{black!70}[1.10,\,1.70]}} & \mbox{1.38\,{\fontsize{7}{8}\selectfont\color{black!70}[1.23,\,1.54]}} & \mbox{1.38\,{\fontsize{7}{8}\selectfont\color{black!70}[1.13,\,1.69]}} & \mbox{1.22\,{\fontsize{7}{8}\selectfont\color{black!70}[1.01,\,1.47]}} & \mbox{1.22\,{\fontsize{7}{8}\selectfont\color{black!70}[1.14,\,1.30]}} & \mbox{1.17\,{\fontsize{7}{8}\selectfont\color{black!70}[1.09,\,1.26]}} \\
\addlinespace[4pt]
\multicolumn{7}{@{}l}{\itshape Backward precision} \\[-1pt]
NVFP4 + FP8 backward & \mbox{1.30\,{\fontsize{7}{8}\selectfont\color{black!70}[1.13,\,1.49]}} & \mbox{1.24\,{\fontsize{7}{8}\selectfont\color{black!70}[1.09,\,1.41]}} & \mbox{1.20\,{\fontsize{7}{8}\selectfont\color{black!70}[1.05,\,1.36]}} & \mbox{1.26\,{\fontsize{7}{8}\selectfont\color{black!70}[1.12,\,1.43]}} & \mbox{1.20\,{\fontsize{7}{8}\selectfont\color{black!70}[1.07,\,1.34]}} & \mbox{1.14\,{\fontsize{7}{8}\selectfont\color{black!70}[1.03,\,1.27]}} \\
Recipe + FP8 backward & \mbox{1.22\,{\fontsize{7}{8}\selectfont\color{black!70}[1.12,\,1.32]}} & \mbox{1.17\,{\fontsize{7}{8}\selectfont\color{black!70}[1.07,\,1.27]}} & \mbox{1.14\,{\fontsize{7}{8}\selectfont\color{black!70}[1.02,\,1.28]}} & \mbox{1.19\,{\fontsize{7}{8}\selectfont\color{black!70}[1.04,\,1.36]}} & \mbox{1.11\,{\fontsize{7}{8}\selectfont\color{black!70}[1.00,\,1.24]}} & \mbox{1.08\,{\fontsize{7}{8}\selectfont\color{black!70}[0.93,\,1.26]}} \\
\addlinespace[4pt]
\multicolumn{7}{@{}l}{\itshape Training settings} \\[-1pt]
Linear learning-rate decay & \mbox{1.46\,{\fontsize{7}{8}\selectfont\color{black!70}[1.24,\,1.72]}} & \mbox{1.44\,{\fontsize{7}{8}\selectfont\color{black!70}[1.27,\,1.63]}} & \mbox{1.38\,{\fontsize{7}{8}\selectfont\color{black!70}[1.08,\,1.78]}} & \mbox{1.34\,{\fontsize{7}{8}\selectfont\color{black!70}[1.24,\,1.45]}} & \mbox{1.27\,{\fontsize{7}{8}\selectfont\color{black!70}[1.09,\,1.47]}} & \mbox{1.17\,{\fontsize{7}{8}\selectfont\color{black!70}[0.90,\,1.51]}} \\
Peak rate $5\times10^{-4}$ & \mbox{1.00\,{\fontsize{7}{8}\selectfont\color{black!70}[0.54,\,1.83]}} & \mbox{0.97\,{\fontsize{7}{8}\selectfont\color{black!70}[0.45,\,2.10]}} & \mbox{0.82\,{\fontsize{7}{8}\selectfont\color{black!70}[0.29,\,2.28]}} & \mbox{0.96\,{\fontsize{7}{8}\selectfont\color{black!70}[0.53,\,1.73]}} & \mbox{0.90\,{\fontsize{7}{8}\selectfont\color{black!70}[0.44,\,1.86]}} & \mbox{0.74\,{\fontsize{7}{8}\selectfont\color{black!70}[0.27,\,2.04]}} \\
Peak rate $2\times10^{-3}$ & \mbox{1.53\,{\fontsize{7}{8}\selectfont\color{black!70}[0.78,\,3.00]}} & \mbox{1.44\,{\fontsize{7}{8}\selectfont\color{black!70}[0.68,\,3.06]}} & \mbox{1.54\,{\fontsize{7}{8}\selectfont\color{black!70}[0.75,\,3.16]}} & \mbox{1.43\,{\fontsize{7}{8}\selectfont\color{black!70}[0.84,\,2.41]}} & \mbox{1.34\,{\fontsize{7}{8}\selectfont\color{black!70}[0.72,\,2.51]}} & \mbox{1.41\,{\fontsize{7}{8}\selectfont\color{black!70}[0.74,\,2.67]}} \\
100M architecture & \mbox{1.38\,{\fontsize{7}{8}\selectfont\color{black!70}[1.20,\,1.57]}} & \mbox{1.32\,{\fontsize{7}{8}\selectfont\color{black!70}[1.13,\,1.55]}} & \mbox{1.28\,{\fontsize{7}{8}\selectfont\color{black!70}[1.18,\,1.40]}} & \mbox{1.31\,{\fontsize{7}{8}\selectfont\color{black!70}[1.12,\,1.55]}} & \mbox{1.24\,{\fontsize{7}{8}\selectfont\color{black!70}[1.03,\,1.49]}} & \mbox{1.17\,{\fontsize{7}{8}\selectfont\color{black!70}[1.04,\,1.32]}} \\
\addlinespace[4pt]
\multicolumn{7}{@{}l}{\itshape Quantization scope} \\[-1pt]
Forward only & \mbox{1.24\,{\fontsize{7}{8}\selectfont\color{black!70}[1.14,\,1.34]}} & \mbox{1.20\,{\fontsize{7}{8}\selectfont\color{black!70}[1.09,\,1.31]}} & \mbox{1.16\,{\fontsize{7}{8}\selectfont\color{black!70}[1.03,\,1.31]}} & \mbox{1.22\,{\fontsize{7}{8}\selectfont\color{black!70}[1.11,\,1.33]}} & \mbox{1.16\,{\fontsize{7}{8}\selectfont\color{black!70}[1.04,\,1.30]}} & \mbox{1.11\,{\fontsize{7}{8}\selectfont\color{black!70}[0.97,\,1.27]}} \\
Forward weights only & \mbox{1.14\,{\fontsize{7}{8}\selectfont\color{black!70}[0.86,\,1.51]}} & \mbox{1.11\,{\fontsize{7}{8}\selectfont\color{black!70}[0.95,\,1.31]}} & \mbox{1.10\,{\fontsize{7}{8}\selectfont\color{black!70}[1.01,\,1.20]}} & \mbox{1.18\,{\fontsize{7}{8}\selectfont\color{black!70}[0.69,\,2.02]}} & \mbox{1.11\,{\fontsize{7}{8}\selectfont\color{black!70}[0.90,\,1.38]}} & \mbox{1.08\,{\fontsize{7}{8}\selectfont\color{black!70}[1.02,\,1.15]}} \\
Forward activations only & \mbox{1.17\,{\fontsize{7}{8}\selectfont\color{black!70}[1.03,\,1.34]}} & \mbox{1.15\,{\fontsize{7}{8}\selectfont\color{black!70}[1.00,\,1.31]}} & \mbox{1.13\,{\fontsize{7}{8}\selectfont\color{black!70}[0.98,\,1.30]}} & \mbox{1.17\,{\fontsize{7}{8}\selectfont\color{black!70}[1.07,\,1.27]}} & \mbox{1.12\,{\fontsize{7}{8}\selectfont\color{black!70}[0.57,\,2.22]}} & \mbox{1.09\,{\fontsize{7}{8}\selectfont\color{black!70}[0.48,\,2.47]}} \\
Backward only & \mbox{1.17\,{\fontsize{7}{8}\selectfont\color{black!70}[1.08,\,1.28]}} & \mbox{1.13\,{\fontsize{7}{8}\selectfont\color{black!70}[1.04,\,1.23]}} & \mbox{1.12\,{\fontsize{7}{8}\selectfont\color{black!70}[1.01,\,1.24]}} & \mbox{1.13\,{\fontsize{7}{8}\selectfont\color{black!70}[1.00,\,1.28]}} & \mbox{1.09\,{\fontsize{7}{8}\selectfont\color{black!70}[0.94,\,1.26]}} & \mbox{1.09\,{\fontsize{7}{8}\selectfont\color{black!70}[0.92,\,1.30]}} \\
\addlinespace[4pt]
\multicolumn{7}{@{}l}{\itshape Noise controls} \\[-1pt]
Forward noise & \mbox{1.26\,{\fontsize{7}{8}\selectfont\color{black!70}[1.15,\,1.39]}} & \mbox{1.20\,{\fontsize{7}{8}\selectfont\color{black!70}[1.08,\,1.34]}} & \mbox{1.16\,{\fontsize{7}{8}\selectfont\color{black!70}[1.03,\,1.30]}} & \mbox{1.27\,{\fontsize{7}{8}\selectfont\color{black!70}[1.14,\,1.43]}} & \mbox{1.19\,{\fontsize{7}{8}\selectfont\color{black!70}[1.01,\,1.41]}} & \mbox{1.12\,{\fontsize{7}{8}\selectfont\color{black!70}[0.94,\,1.33]}} \\
i.i.d. gradient noise & \mbox{1.27\,{\fontsize{7}{8}\selectfont\color{black!70}[1.08,\,1.48]}} & \mbox{1.26\,{\fontsize{7}{8}\selectfont\color{black!70}[1.09,\,1.45]}} & \mbox{1.24\,{\fontsize{7}{8}\selectfont\color{black!70}[1.08,\,1.42]}} & \mbox{1.28\,{\fontsize{7}{8}\selectfont\color{black!70}[1.12,\,1.46]}} & \mbox{1.25\,{\fontsize{7}{8}\selectfont\color{black!70}[1.10,\,1.42]}} & \mbox{1.20\,{\fontsize{7}{8}\selectfont\color{black!70}[1.06,\,1.37]}} \\
Block-shaped gradient noise & \mbox{1.15\,{\fontsize{7}{8}\selectfont\color{black!70}[1.03,\,1.29]}} & \mbox{1.12\,{\fontsize{7}{8}\selectfont\color{black!70}[1.03,\,1.22]}} & \mbox{1.11\,{\fontsize{7}{8}\selectfont\color{black!70}[1.01,\,1.22]}} & \mbox{1.16\,{\fontsize{7}{8}\selectfont\color{black!70}[1.01,\,1.32]}} & \mbox{1.12\,{\fontsize{7}{8}\selectfont\color{black!70}[1.01,\,1.24]}} & \mbox{1.10\,{\fontsize{7}{8}\selectfont\color{black!70}[0.97,\,1.24]}} \\
\bottomrule
\end{tabular*}
\end{table}
\begin{table}[tbp]
\centering
\caption{Sensitivity to the recall readout (value prob.: correct-value probability). BF16-forward values of NVFP4 for seeds 0, 1, 2 come from a rerun of the same configuration with saved checkpoints. Geometric means with 95\% intervals. Seed counts are as in Table~\ref{tab:robust}, except where a cell ends with its own \emph{n}. --: unavailable or outside the BF16 loss range.}
\label{tab:robust-readouts}
\footnotesize
\setlength{\tabcolsep}{2pt}
\renewcommand{\arraystretch}{1.0}
\begin{tabular*}{\textwidth}{@{\extracolsep{\fill}}lrrrrrr@{}}
\toprule
Setting & \multicolumn{3}{c}{Matched tokens} & \multicolumn{3}{c}{Matched loss} \\
\cmidrule(lr){2-4}\cmidrule(l){5-7}
 & Exact match & BF16 forward & Value prob. & Exact match & BF16 forward & Value prob. \\
\midrule
\addlinespace[4pt]
\multicolumn{7}{@{}l}{\itshape Formats} \\[-1pt]
FP8 & \mbox{1.10\,{\fontsize{7}{8}\selectfont\color{black!70}[1.00,\,1.20]}} & \mbox{1.10\,{\fontsize{7}{8}\selectfont\color{black!70}[1.00,\,1.21]}} & \mbox{1.14\,{\fontsize{7}{8}\selectfont\color{black!70}[1.02,\,1.28]}} & \mbox{1.09\,{\fontsize{7}{8}\selectfont\color{black!70}[0.95,\,1.25]}} & \mbox{1.10\,{\fontsize{7}{8}\selectfont\color{black!70}[0.96,\,1.26]}} & \mbox{1.10\,{\fontsize{7}{8}\selectfont\color{black!70}[1.00,\,1.21]}} \\
NVFP4 & \mbox{1.31\,{\fontsize{7}{8}\selectfont\color{black!70}[1.16,\,1.48]}} & \mbox{1.26\,{\fontsize{7}{8}\selectfont\color{black!70}[1.11,\,1.44]}} & \mbox{1.32\,{\fontsize{7}{8}\selectfont\color{black!70}[1.06,\,1.64]}} & \mbox{1.24\,{\fontsize{7}{8}\selectfont\color{black!70}[1.13,\,1.35]}} & \mbox{1.21\,{\fontsize{7}{8}\selectfont\color{black!70}[1.10,\,1.34]}} & \mbox{1.19\,{\fontsize{7}{8}\selectfont\color{black!70}[0.99,\,1.43]}} \\
MXFP4 & \mbox{1.62\,{\fontsize{7}{8}\selectfont\color{black!70}[1.38,\,1.89]}} & \mbox{1.47\,{\fontsize{7}{8}\selectfont\color{black!70}[1.29,\,1.69]}} & \mbox{1.71\,{\fontsize{7}{8}\selectfont\color{black!70}[1.37,\,2.13]}} & \mbox{1.42\,{\fontsize{7}{8}\selectfont\color{black!70}[1.22,\,1.64]}} & \mbox{1.32\,{\fontsize{7}{8}\selectfont\color{black!70}[1.17,\,1.50]}} & \mbox{1.34\,{\fontsize{7}{8}\selectfont\color{black!70}[1.08,\,1.65]}} \\
\addlinespace[4pt]
\multicolumn{7}{@{}l}{\itshape Recipes} \\[-1pt]
NVIDIA recipe (adapted) & \mbox{1.17\,{\fontsize{7}{8}\selectfont\color{black!70}[1.09,\,1.25]}} & \mbox{1.15\,{\fontsize{7}{8}\selectfont\color{black!70}[1.07,\,1.24]}} & \mbox{1.20\,{\fontsize{7}{8}\selectfont\color{black!70}[1.05,\,1.38]}} & \mbox{1.15\,{\fontsize{7}{8}\selectfont\color{black!70}[1.08,\,1.22]}} & \mbox{1.15\,{\fontsize{7}{8}\selectfont\color{black!70}[1.07,\,1.24]}\,{\scriptsize\itshape n=5}} & \mbox{1.15\,{\fontsize{7}{8}\selectfont\color{black!70}[1.00,\,1.32]}} \\
Simplified recipe & \mbox{1.20\,{\fontsize{7}{8}\selectfont\color{black!70}[1.09,\,1.31]}} & \mbox{1.17\,{\fontsize{7}{8}\selectfont\color{black!70}[1.06,\,1.30]}} & \mbox{1.21\,{\fontsize{7}{8}\selectfont\color{black!70}[1.03,\,1.42]}} & \mbox{1.16\,{\fontsize{7}{8}\selectfont\color{black!70}[1.05,\,1.27]}} & \mbox{1.14\,{\fontsize{7}{8}\selectfont\color{black!70}[1.03,\,1.27]}} & \mbox{1.13\,{\fontsize{7}{8}\selectfont\color{black!70}[0.95,\,1.35]}} \\
BF16 cooldown & \mbox{1.19\,{\fontsize{7}{8}\selectfont\color{black!70}[1.07,\,1.32]}} & \mbox{1.20\,{\fontsize{7}{8}\selectfont\color{black!70}[1.08,\,1.33]}} & \mbox{1.18\,{\fontsize{7}{8}\selectfont\color{black!70}[1.00,\,1.40]}} & \mbox{1.17\,{\fontsize{7}{8}\selectfont\color{black!70}[1.08,\,1.27]}} & \mbox{1.18\,{\fontsize{7}{8}\selectfont\color{black!70}[1.09,\,1.29]}} & \mbox{1.13\,{\fontsize{7}{8}\selectfont\color{black!70}[0.96,\,1.33]}} \\
\addlinespace[4pt]
\multicolumn{7}{@{}l}{\itshape Natural-text mixture} \\[-1pt]
NVFP4 & \mbox{1.51\,{\fontsize{7}{8}\selectfont\color{black!70}[1.37,\,1.68]}} & \mbox{1.30\,{\fontsize{7}{8}\selectfont\color{black!70}[1.13,\,1.50]}} & \mbox{1.46\,{\fontsize{7}{8}\selectfont\color{black!70}[1.24,\,1.72]}} & \mbox{1.19\,{\fontsize{7}{8}\selectfont\color{black!70}[1.06,\,1.34]}} & \mbox{1.11\,{\fontsize{7}{8}\selectfont\color{black!70}[0.97,\,1.26]}} & \mbox{1.09\,{\fontsize{7}{8}\selectfont\color{black!70}[0.90,\,1.31]}} \\
NVIDIA recipe (adapted) & \mbox{1.36\,{\fontsize{7}{8}\selectfont\color{black!70}[1.21,\,1.53]}} & \mbox{1.25\,{\fontsize{7}{8}\selectfont\color{black!70}[1.02,\,1.54]}} & \mbox{1.26\,{\fontsize{7}{8}\selectfont\color{black!70}[1.04,\,1.53]}} & \mbox{1.16\,{\fontsize{7}{8}\selectfont\color{black!70}[1.13,\,1.20]}} & \mbox{1.12\,{\fontsize{7}{8}\selectfont\color{black!70}[0.96,\,1.30]}} & \mbox{1.03\,{\fontsize{7}{8}\selectfont\color{black!70}[0.87,\,1.22]}} \\
\addlinespace[4pt]
\multicolumn{7}{@{}l}{\itshape Backward precision} \\[-1pt]
NVFP4 + FP8 backward & \mbox{1.22\,{\fontsize{7}{8}\selectfont\color{black!70}[1.09,\,1.36]}} & \mbox{1.18\,{\fontsize{7}{8}\selectfont\color{black!70}[1.04,\,1.33]}} & \mbox{1.23\,{\fontsize{7}{8}\selectfont\color{black!70}[1.08,\,1.40]}} & \mbox{1.18\,{\fontsize{7}{8}\selectfont\color{black!70}[1.07,\,1.29]}} & \mbox{1.17\,{\fontsize{7}{8}\selectfont\color{black!70}[1.03,\,1.34]}\,{\scriptsize\itshape n=5}} & \mbox{1.15\,{\fontsize{7}{8}\selectfont\color{black!70}[1.04,\,1.27]}} \\
Recipe + FP8 backward & \mbox{1.15\,{\fontsize{7}{8}\selectfont\color{black!70}[1.06,\,1.26]}} & \mbox{1.13\,{\fontsize{7}{8}\selectfont\color{black!70}[1.03,\,1.25]}} & \mbox{1.12\,{\fontsize{7}{8}\selectfont\color{black!70}[0.93,\,1.35]}} & \mbox{1.11\,{\fontsize{7}{8}\selectfont\color{black!70}[0.98,\,1.24]}} & \mbox{1.15\,{\fontsize{7}{8}\selectfont\color{black!70}[0.99,\,1.33]}\,{\scriptsize\itshape n=3}} & \mbox{1.04\,{\fontsize{7}{8}\selectfont\color{black!70}[0.78,\,1.38]}} \\
\addlinespace[4pt]
\multicolumn{7}{@{}l}{\itshape Training settings} \\[-1pt]
Linear learning-rate decay & \mbox{1.39\,{\fontsize{7}{8}\selectfont\color{black!70}[1.19,\,1.62]}} & \mbox{1.32\,{\fontsize{7}{8}\selectfont\color{black!70}[1.19,\,1.47]}} & \mbox{1.30\,{\fontsize{7}{8}\selectfont\color{black!70}[1.06,\,1.59]}} & \mbox{1.24\,{\fontsize{7}{8}\selectfont\color{black!70}[1.07,\,1.43]}} & \mbox{1.23\,{\fontsize{7}{8}\selectfont\color{black!70}[1.09,\,1.39]}} & \mbox{1.03\,{\fontsize{7}{8}\selectfont\color{black!70}[0.85,\,1.26]}} \\
Peak rate $5\times10^{-4}$ & \mbox{0.89\,{\fontsize{7}{8}\selectfont\color{black!70}[0.37,\,2.16]}} & \mbox{0.83\,{\fontsize{7}{8}\selectfont\color{black!70}[0.35,\,2.01]}} & \mbox{0.58\,{\fontsize{7}{8}\selectfont\color{black!70}[0.15,\,2.23]}} & \mbox{0.82\,{\fontsize{7}{8}\selectfont\color{black!70}[0.35,\,1.92]}} & \mbox{0.80\,{\fontsize{7}{8}\selectfont\color{black!70}[0.34,\,1.89]}} & \mbox{0.50\,{\fontsize{7}{8}\selectfont\color{black!70}[0.14,\,1.82]}} \\
Peak rate $2\times10^{-3}$ & \mbox{1.49\,{\fontsize{7}{8}\selectfont\color{black!70}[0.72,\,3.10]}} & \mbox{1.42\,{\fontsize{7}{8}\selectfont\color{black!70}[0.66,\,3.07]}} & \mbox{1.56\,{\fontsize{7}{8}\selectfont\color{black!70}[0.71,\,3.41]}} & \mbox{1.37\,{\fontsize{7}{8}\selectfont\color{black!70}[0.76,\,2.47]}} & \mbox{1.34\,{\fontsize{7}{8}\selectfont\color{black!70}[0.72,\,2.49]}} & \mbox{1.34\,{\fontsize{7}{8}\selectfont\color{black!70}[0.83,\,2.16]}} \\
100M architecture & \mbox{1.32\,{\fontsize{7}{8}\selectfont\color{black!70}[1.16,\,1.51]}} & \mbox{1.24\,{\fontsize{7}{8}\selectfont\color{black!70}[1.14,\,1.36]}} & \mbox{1.32\,{\fontsize{7}{8}\selectfont\color{black!70}[1.21,\,1.43]}} & \mbox{1.24\,{\fontsize{7}{8}\selectfont\color{black!70}[1.07,\,1.43]}} & \mbox{1.19\,{\fontsize{7}{8}\selectfont\color{black!70}[1.05,\,1.34]}} & \mbox{1.15\,{\fontsize{7}{8}\selectfont\color{black!70}[1.11,\,1.20]}} \\
\addlinespace[4pt]
\multicolumn{7}{@{}l}{\itshape Quantization scope} \\[-1pt]
Forward only & \mbox{1.18\,{\fontsize{7}{8}\selectfont\color{black!70}[1.07,\,1.30]}} & \mbox{1.13\,{\fontsize{7}{8}\selectfont\color{black!70}[1.04,\,1.24]}} & \mbox{1.16\,{\fontsize{7}{8}\selectfont\color{black!70}[0.99,\,1.36]}} & \mbox{1.15\,{\fontsize{7}{8}\selectfont\color{black!70}[1.04,\,1.28]}} & \mbox{1.12\,{\fontsize{7}{8}\selectfont\color{black!70}[1.02,\,1.24]}} & \mbox{1.09\,{\fontsize{7}{8}\selectfont\color{black!70}[0.92,\,1.31]}} \\
Forward weights only & \mbox{1.10\,{\fontsize{7}{8}\selectfont\color{black!70}[0.97,\,1.25]}} & \mbox{1.10\,{\fontsize{7}{8}\selectfont\color{black!70}[0.93,\,1.29]}} & \mbox{1.14\,{\fontsize{7}{8}\selectfont\color{black!70}[0.92,\,1.42]}} & \mbox{1.10\,{\fontsize{7}{8}\selectfont\color{black!70}[0.93,\,1.30]}} & \mbox{1.11\,{\fontsize{7}{8}\selectfont\color{black!70}[0.98,\,1.26]}} & \mbox{1.14\,{\fontsize{7}{8}\selectfont\color{black!70}[0.98,\,1.33]}} \\
Forward activations only & \mbox{1.13\,{\fontsize{7}{8}\selectfont\color{black!70}[1.01,\,1.26]}} & \mbox{1.11\,{\fontsize{7}{8}\selectfont\color{black!70}[0.99,\,1.23]}} & \mbox{1.14\,{\fontsize{7}{8}\selectfont\color{black!70}[0.92,\,1.41]}} & \mbox{1.10\,{\fontsize{7}{8}\selectfont\color{black!70}[0.62,\,1.95]}} & \mbox{1.09\,{\fontsize{7}{8}\selectfont\color{black!70}[0.64,\,1.88]}} & \mbox{1.06\,{\fontsize{7}{8}\selectfont\color{black!70}[0.29,\,3.89]}} \\
Backward only & \mbox{1.11\,{\fontsize{7}{8}\selectfont\color{black!70}[1.03,\,1.19]}} & \mbox{1.13\,{\fontsize{7}{8}\selectfont\color{black!70}[1.04,\,1.21]}} & \mbox{1.14\,{\fontsize{7}{8}\selectfont\color{black!70}[1.08,\,1.21]}} & \mbox{1.07\,{\fontsize{7}{8}\selectfont\color{black!70}[0.96,\,1.20]}} & \mbox{1.09\,{\fontsize{7}{8}\selectfont\color{black!70}[0.96,\,1.24]}} & \mbox{1.09\,{\fontsize{7}{8}\selectfont\color{black!70}[1.00,\,1.19]}} \\
\addlinespace[4pt]
\multicolumn{7}{@{}l}{\itshape Noise controls} \\[-1pt]
Forward noise & \mbox{1.18\,{\fontsize{7}{8}\selectfont\color{black!70}[1.08,\,1.30]}} & \mbox{1.19\,{\fontsize{7}{8}\selectfont\color{black!70}[1.08,\,1.31]}} & \mbox{1.13\,{\fontsize{7}{8}\selectfont\color{black!70}[1.03,\,1.25]}} & \mbox{1.17\,{\fontsize{7}{8}\selectfont\color{black!70}[1.02,\,1.34]}} & \mbox{1.18\,{\fontsize{7}{8}\selectfont\color{black!70}[1.03,\,1.36]}} & \mbox{1.06\,{\fontsize{7}{8}\selectfont\color{black!70}[0.92,\,1.22]}} \\
i.i.d. gradient noise & \mbox{1.24\,{\fontsize{7}{8}\selectfont\color{black!70}[1.08,\,1.42]}} & \mbox{1.25\,{\fontsize{7}{8}\selectfont\color{black!70}[1.09,\,1.44]}} & \mbox{1.36\,{\fontsize{7}{8}\selectfont\color{black!70}[1.18,\,1.58]}} & \mbox{1.23\,{\fontsize{7}{8}\selectfont\color{black!70}[1.09,\,1.39]}} & \mbox{1.24\,{\fontsize{7}{8}\selectfont\color{black!70}[1.09,\,1.41]}} & \mbox{1.31\,{\fontsize{7}{8}\selectfont\color{black!70}[1.20,\,1.44]}} \\
Block-shaped gradient noise & \mbox{1.10\,{\fontsize{7}{8}\selectfont\color{black!70}[1.01,\,1.20]}} & \mbox{1.11\,{\fontsize{7}{8}\selectfont\color{black!70}[1.03,\,1.21]}} & \mbox{1.13\,{\fontsize{7}{8}\selectfont\color{black!70}[1.02,\,1.25]}} & \mbox{1.10\,{\fontsize{7}{8}\selectfont\color{black!70}[0.99,\,1.22]}} & \mbox{1.11\,{\fontsize{7}{8}\selectfont\color{black!70}[1.01,\,1.22]}} & \mbox{1.10\,{\fontsize{7}{8}\selectfont\color{black!70}[0.97,\,1.26]}} \\
\bottomrule
\end{tabular*}
\end{table}

Table~\ref{tab:robust} reports the default multipliers, and Tables~\ref{tab:robust-levels}
and~\ref{tab:robust-readouts} separate sensitivity to the recall level from sensitivity to the readout. \emph{Recall levels}: the main
multiplier averages the $\log_2$-exposure gap over recall levels 20--80\%. On synthetic data it is larger at 30\% than
at 70\% in 18 of 19 settings, typically by a quarter on the $\log_2$ scale. On natural text the three levels agree
(Section~\ref{sec:main}). To check that this level dependence is not produced by the coarse exposure grid, we interpolate each
BF16 recall curve by a monotone cubic, shift it by the measured \nvfp{} multiplier, and read both curves on the same
grid as the data. The ratio of the $\log_2$ shifts at 70\% and 30\% is then 0.93 on synthetic data. \emph{Exact match}
of the whole value instead of the first value token moves no multiplier in Figure~\ref{fig:mult} by more than 0.07.
\emph{BF16-forward evaluation} reads the low-precision weights with BF16 forward GEMMs, removing quantization error at
evaluation time, and also uses the BF16-forward validation loss for loss matching. It is available for all
natural-text seeds and for synthetic seeds 3--5 of the original runs. For synthetic seeds 0--2 it comes from reruns of
the same configuration with saved checkpoints, whose default-readout multipliers agree with the original runs to within 0.03.
The token-matched cost remains (1.26$\times$ [1.11, 1.44] synthetic, 1.30$\times$ [1.13, 1.50] natural text), and at
matched loss it is 1.21$\times$ [1.10, 1.34] on synthetic data and shrinks to 1.11$\times$ [0.97, 1.26] on natural
text. \emph{Correct-value probability} replaces first-token accuracy by the probability the model
assigns to the whole correct value. It gives the same token-matched multipliers (1.32$\times$
synthetic, 1.46$\times$ natural text) but noisier loss-matched ones, whose intervals include 1 for plain \nvfp{}
(1.19$\times$, 1.09$\times$) and for the adapted recipe on natural text (1.03$\times$). For plain \nvfp{}, token-matched
estimates therefore exceed 1 under every tested readout. The loss-matched point estimate exceeds 1 at every recall
level and under exact match, with an interval that includes 1 only at 70\% on natural text, but part of it disappears
under BF16-forward evaluation on natural text and under correct-value probability. \emph{Fixed facts}: the probed facts
are the same in every seed, so the seed intervals do not include uncertainty from the choice of facts. Resampling the
200 probed persons of each bin together with the seeds (2{,}000 draws, matched tokens, per-probe records of five
synthetic and six natural-text seeds) gives [1.20, 1.49] for synthetic \nvfp{} against [1.20, 1.48] when only the
seeds are resampled, and [1.40, 1.66] against [1.41, 1.65] on natural text. The choice of facts therefore adds little
uncertainty.

\section{Additional Controls}\label{app:controls}
\paragraph{Recipes at matched loss.}
Table~\ref{tab:recipe} lists loss-matched multipliers for Section~\ref{sec:recipe} and paired tests against plain
\nvfp{}.
\begin{table}[h]
\centering
\caption{Recipe variants of plain \nvfp{} (40M synthetic). $m$: loss-matched multiplier (six seeds, five for BF16
cooldown, four for the recipe with FP8 backward). $p$: paired $t$-test of $\log_2 m$ against plain \nvfp{}. After
Holm's correction over the five tests, the two smallest $p$ remain below 0.05.}
\label{tab:recipe}
\small
\renewcommand{\arraystretch}{1.18}
\begin{tabularx}{0.80\textwidth}{@{}>{\raggedright\arraybackslash}Xrrr@{}}
\toprule
Variant & Loss gap & $m$ [95\% CI] & $p$ \\
\midrule
Plain \nvfp{} & 0.47\% & 1.25 [1.13, 1.38] & -- \\
Simplified recipe (BF16 last 2 blocks, RHT also on data gradient) & 0.27\% & 1.16 [1.04, 1.29] & 0.17 \\
BF16 cooldown (final 20\%) & 0.25\% & 1.18 [1.09, 1.29] & 0.02 \\
NVIDIA recipe (adapted) & 0.19\% & 1.15 [1.07, 1.23] & 0.008 \\
\nvfp{} with FP8 backward GEMMs & 0.29\% & 1.19 [1.07, 1.33] & 0.46 \\
NVIDIA recipe (adapted) with FP8 backward GEMMs & 0.19\% & 1.12 [0.98, 1.27] & 0.008 \\
\bottomrule
\end{tabularx}
\end{table}

\paragraph{Recovery shares.}
For a remedy compared with plain \nvfp{} on the $n$ seeds where both have a defined token-matched multiplier, let
$a_j$ and $b_j$ be the $\log_2$ multipliers of plain \nvfp{} and of the remedy on seed $j$, and $\lambda_j$ and
$\kappa_j$ their relative validation-loss gaps to the BF16 run of that seed. With bars denoting means over the $n$
seeds, Figure~\ref{fig:recovery} plots the cost share $C=1-\bar b/\bar a$ against the loss share
$1-\bar\kappa/\bar\lambda$. The interval for $C$ comes from the Fieller interval for the ratio of the expected values of $b$ and $a$, the set of $\theta$
with $(\bar b-\theta\bar a)^2\le t^2_{n-1,0.975}\,(S_{bb}-2\theta S_{ab}+\theta^2 S_{aa})/n$, where $S_{aa}$, $S_{bb}$
and $S_{ab}$ are the sample variances and covariance over seeds and $t_{n-1,0.975}$ is the 97.5th percentile of the
$t$ distribution with $n-1$ degrees of freedom. The loss-share interval uses $\lambda_j$ and $\kappa_j$ in place of
$a_j$ and $b_j$. For the difference between the two shares we resample
seeds with replacement 20{,}000 times, recompute both shares from the same resampled seeds, and report the 2.5th and
97.5th percentiles. Table~\ref{tab:recovery} gives both shares with Fieller intervals and their difference.

\begin{table}[h]
\centering
\caption{Shares of the validation-loss gap and of the $\log_2$ exposure cost (matched tokens) of plain \nvfp{} that each remedy removes, in percent, with 95\% Fieller intervals, and their difference (loss share minus cost share, in percentage points) with a 95\% seed-bootstrap interval. --: unbounded interval, or too few seeds for a bootstrap.}
\label{tab:recovery}
\small
\begin{tabular}{@{}lrrrr@{}}
\toprule
Remedy & $n$ & Loss share & Cost share & Difference \\
\midrule
NVIDIA recipe (adapted) & 6 & 60 [29, 79] & 42 [26, 56] & 18 [2, 29] \\
Simplified recipe & 6 & 42 [$-$18, 59] & 34 [$-$19, 70] & 7 [$-$26, 37] \\
BF16 cooldown & 6 & 46 [31, 105] & 35 [23, 53] & 11 [3, 29] \\
FP8 backward & 6 & 39 [8, 67] & 25 [$-$56, 70] & 14 [$-$4, 42] \\
Recipe + FP8 backward & 6 & 58 [29, 131] & 46 [32, 65] & 13 [$-$10, 49] \\
Adapted recipe, natural & 3 & 41 [32, 51] & 12 [$-$75, 45] & 29 [--] \\
\bottomrule
\end{tabular}
\end{table}

\begin{figure}[h]
\centering
\includegraphics[width=0.48\textwidth]{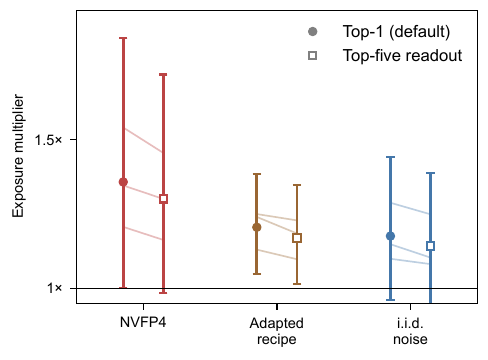}
\caption{Exposure multiplier at matched tokens (40M, seeds 0--2, 95\% intervals) under the default top-1 readout
(filled) and a top-five readout over the values of the attribute (open). Thin lines pair the two readouts of one
seed.}
\label{fig:probe}
\end{figure}

\paragraph{Lenient readout.}
Figure~\ref{fig:probe} shows the per-seed comparison of top-1 and top-five recall summarized in
Section~\ref{sec:mech}.

\paragraph{Intermediate-layer readouts and probe limitations.}
Under the logit-lens readout, rare facts become accessible in the last three of eight blocks. BF16 reads 18\%, 49\%
and 83\% of the facts seen 4--16 times after blocks 6, 7 and 8. The per-layer deficit varies too much across seeds
to localize it further. We do not use trained linear probes~\citep{allenzhu2023physics31}. With 200--300 values per
attribute and 20k persons, a probe on the name representation stayed at chance for every model. On a small control
model, a probe at the pre-value position recovered only about half of the facts the model generates.

\paragraph{Shared learning rates.}
On seeds 0--2, the token-matched multiplier at a shared rate is 0.91$\times$ [0.38, 2.16] at $5\times10^{-4}$,
1.37$\times$ [0.92, 2.04] at $10^{-3}$, and 1.49$\times$ [0.70, 3.18] at $2\times10^{-3}$. Wide intervals prevent a precise
comparison of the endpoints. At $5\times10^{-4}$, BF16 recalls 58--84\% of facts seen eight times across seeds,
versus 77--78\% for \nvfp{}, limiting resolution at the lower rate.
Across seeds 0--2, mean recall over the 1--8-exposure bins at rates $5\times10^{-4}$, $10^{-3}$ and $2\times10^{-3}$
is 0.371/0.498/0.339 for BF16 and 0.378/0.403/0.234 for \nvfp{}. A quadratic in $\log_2$ of the rate through these
three points peaks at about 0.42 for \nvfp{} (near $8\times10^{-4}$) and 0.50 for BF16. This three-point interpolation
is descriptive, and it gives no indication of a rate within the tested range that closes the gap. Both formats select
$10^{-3}$ under this criterion, where the three-seed loss-matched multiplier is 1.30$\times$ [0.98, 1.71]. Comparing \nvfp{} at $5\times10^{-4}$ with
BF16 at $10^{-3}$ instead gives 1.47$\times$ [1.42, 1.53] at matched tokens and 1.40$\times$ [1.31, 1.49] at matched
loss. These comparisons use the same three seeds, but the latter selects a different \nvfp{} rate.

\paragraph{BF16 cooldown by exposure timing.}
The cooldown run switches from \nvfp{} to BF16 at step 1{,}590 of 1{,}988. From the data-stream index we take every
probed fact of the bins seen 1--8 times and classify it by the steps of its exposures, either all before the switch
or all after it. For each group and seed, the deficit of a run is the mean recall of BF16 minus that of the run over
the group's facts. Plain \nvfp{} is evaluated with BF16 matrix multiplications (seeds 0--2 from reruns that store this
readout), so that it is read in the same arithmetic as the cooldown run. We report the mean recall difference between
the cooldown run and plain \nvfp{} with a $t$ interval over the six seeds, and the remaining share, the mean cooldown
deficit over the mean \nvfp{} deficit, with an interval from 20{,}000 seed resamples.

\section{Proofs and a Linearized Model}\label{app:proof}
\subsection{\texorpdfstring{Proof of Proposition~\ref{prop:shift}}{Proof of Proposition 1}}\label{app:proof-shift}
\begin{proof}
(i) For every $\ell$,
$x_{R_{\mathrm{lo}}}(\ell)=\inf\{x: R_{\mathrm{ref}}(x-\delta)\ge\ell\}=\delta+\inf\{y: R_{\mathrm{ref}}(y)\ge\ell\}
=\delta+x_{R_{\mathrm{ref}}}(\ell)$, with $y=x-\delta$. Hence $x_{R_{\mathrm{lo}}}(\ell)$ is finite exactly when
$x_{R_{\mathrm{ref}}}(\ell)$ is, and $\Delta_\ell=\delta$ for every $\ell\in L'$, $\bar\Delta=\delta$ and $m=2^\delta$ whenever $L'$ is nonempty.
(ii) Since $h$ is strictly increasing, $h(R(x))\ge h(\ell)$ if and only if $R(x)\ge\ell$, so the two sets in the
definition of $x_{h\circ R}(h(\ell))$ and $x_R(\ell)$ coincide and $x_{h\circ R}(h(\ell))=x_R(\ell)$ for both curves.
Hence the gap at level $h(\ell)$ equals $\Delta_\ell$ and is finite exactly when $\Delta_\ell$ is, so $L'$ maps to
$h(L')$, and $\bar\Delta$ and $m$ are unchanged. The running maximum commutes with $h$, so the
statement also holds for curves made nondecreasing after the transformation.
\end{proof}
On the observed grid $x\in\{0,\dots,8\}$ with linear interpolation, (i) holds exactly for integer $\delta$ when
levels are bracketed by both curves, and up to interpolation error otherwise. Part (ii) holds for the interpolated
curves only for affine $h$, since a nonlinear $h$ does not preserve linear interpolation. Both are therefore
properties of the measure rather than guarantees about the finite-grid estimate. Table~\ref{tab:robust-levels} and
Appendix~\ref{app:robust} examine (i) empirically. The alternative readouts of Table~\ref{tab:robust-readouts} are not
monotone transforms of first-token recall, so (ii) does not cover them.

\subsection{\texorpdfstring{Proof of Proposition~\ref{prop:adam}}{Proof of Proposition 2}}\label{app:proof-adam}
\begin{proof}
With learning rate $\eta$ and Adam's step $-\eta\,\mu_i/\sqrt{\hat v_i}$, where $\mu_i$ is the momentum (Adam's
first-moment estimate), an exposure of $f$ adds $g_f$ to the
minibatch gradient (up to the batch normalization, a constant factor that cancels in $m_f$) and moves coordinate $i$ by $-\eta\,g_{f,i}/\sqrt{\hat v_i}$ in expectation: the momentum average
spreads the step over later updates, but its weights sum to one, and $\hat v_i$ changes little over that horizon. By
Assumption~\ref{ass:acc}(a) the error adds nothing to this expectation and changes only the normalizer, so the
first-order decrease of the fact's loss per exposure is $P_f=\eta\sum_i g_{f,i}^2/\sqrt{\hat v_i}$, with
$\hat v_i=v_i$ for the reference and $\hat v_i=v_i+s_i^2$ for the low-precision configuration. By
Assumption~\ref{ass:acc}(b) the number of exposures needed is $\tau_f/P_f$, so the ratio between the two is
$P_{f,\mathrm{ref}}/P_{f,\mathrm{lo}}=m_f$, whatever $\tau_f$. Every term of the denominator of Eq.~\eqref{eq:mf} is
at most the matching term of the numerator, so $m_f\ge1$. If $s_i^2=rv_i$, every term is the matching one divided by
$\sqrt{1+r}$, and replacing $s_i^2$ by $c^2s_i^2$ replaces $r$ by $c^2r$. For general $s_i^2$, write
$\rho_i=s_i^2/v_i$ and $w_i=g_{f,i}^2/\sqrt{v_i}$ and neglect $\epsilon$. Then
$(1+c^2\rho_i)^{-1/2}=1-c^2\rho_i/2+O(c^4)$, so the denominator of
Eq.~\eqref{eq:mf} is $(\sum_iw_i)(1-c^2\bar\rho_f/2)+O(c^4)$ with $\bar\rho_f=\sum_iw_i\rho_i/\sum_iw_i$, and
$m_f(c)=1+c^2\bar\rho_f/2+O(c^4)$. For (P3), an error with $s_i^2=\sigma^2$ on every coordinate gives
$\bar\rho_f=\sigma^2\langle 1/v\rangle_w$, and one with the same total power and $s_i^2=\sigma^2v_i/\langle v\rangle$
gives $\bar\rho_f=\sigma^2/\langle v\rangle$, so the first is larger if and only if
$\langle 1/v\rangle_w>1/\langle v\rangle$.
Jensen's inequality gives $\langle 1/v\rangle\ge1/\langle v\rangle$, strictly unless all $v_i$ are equal, which
settles uniform weights. In general $\langle 1/v\rangle_w=\langle w/v\rangle/\langle w\rangle$, which is at least
$\langle 1/v\rangle$ whenever $w$ and $1/v$ are nonnegatively correlated across coordinates. If $g_{f,i}^2$ is
uncorrelated across coordinates with $v_i^{-1/2}$ and $v_i^{-3/2}$, as when it is independent of $v_i$ and the
coordinates are many, then $\langle 1/v\rangle_w=\langle v^{-3/2}\rangle/\langle v^{-1/2}\rangle\ge
\langle 1/v\rangle$, because $v^{-1/2}$ and $v^{-1}$ are both decreasing in $v$ and hence nonnegatively correlated
(Chebyshev's sum inequality). Otherwise the
condition is a property of the facts that must be checked empirically.

For (ii) and (iii), write $g_i=\mathbb{E}[g_i]+\xi_i$ with $\mathrm{Var}(\xi_i)=v_i-\mathbb{E}[g_i]^2$. The momentum is a
weighted sum of past gradients with weights summing to one, so at fixed normalizer the update of coordinate $i$ has
mean $-\eta\,\mathbb{E}[g_i]/\sqrt{\hat v_i}$, and, because the noise and the error are uncorrelated across steps, its
zero-mean part has variance $\kappa\,\eta^2\,\mathrm{Var}(\xi_i+e_i)/\hat v_i$ with the same factor $\kappa$ (set by
$\beta_1$) in every configuration. With the error in both moments, the mean is divided by $\sqrt{1+\rho_i}$ and the variance is
multiplied by $(v_i-\mathbb{E}[g_i]^2+s_i^2)\,v_i/\big((v_i+s_i^2)(v_i-\mathbb{E}[g_i]^2)\big)$, so the ratio of mean to
standard deviation falls by exactly $\sqrt{1+s_i^2/(v_i-\mathbb{E}[g_i]^2)}$, which is $\sqrt{1+\rho_i}$ to leading
order when $\mathbb{E}[g_i]^2\ll v_i$. The variance is then unchanged to leading order. With the error only in the
second moment the momentum is the reference one and the normalizer is $\sqrt{v_i}\sqrt{1+\rho_i}$, so the update is the
reference update with $\eta$ replaced by $\eta/\sqrt{1+\rho_i}$. With the error only in the first moment the normalizer
is the reference one, the mean is unchanged and the variance grows by $\mathrm{Var}(\xi_i+e_i)/\mathrm{Var}(\xi_i)=
1+s_i^2/(v_i-\mathbb{E}[g_i]^2)$.
\end{proof}

\subsection{\texorpdfstring{Proof of Proposition~\ref{prop:loss}}{Proof of Proposition 3}}\label{app:proof-loss}
\begin{proof}
A fact with log-exposure $x$ contributes to the validation stream in proportion to $2^x$, so the excess loss averaged
over fact occurrences in the stream is the first expression. The mean value theorem gives
$\mathcal{L}(x-\delta)-\mathcal{L}(x)=-\delta\,\mathcal{L}'(\xi_x)$ for some $\xi_x\in[x-\delta,x]$ in each term, which
gives the second. If recall is a fixed decreasing function of the value-token loss, the recall curves inherit the
shift by $\delta$, and Proposition~\ref{prop:shift}(i) gives $\Delta_\ell=\delta$ for every $\ell\in L'$ and $m=2^\delta$.
\end{proof}

\subsection{Linearized Dynamics of One Coordinate}\label{app:linear}
To see how the step-size and signal-to-noise effects differ over time, let the mean gradient of coordinate $i$ near
its optimum be $h\,(\theta-\theta^*)$, add zero-mean noise $\xi_t$ of variance $\sigma^2$ and error $e_t$ of variance
$s^2$, both uncorrelated across steps, and hold the normalizer $\sqrt{\hat v}$ fixed. Ignoring momentum, the update
$\theta_{t+1}-\theta^*=(1-k)(\theta_t-\theta^*)-\eta(\xi_t+e_t)/\sqrt{\hat v}$ with $k=\eta h/\sqrt{\hat v}\ll1$ is a
discretized Ornstein--Uhlenbeck process. An exposure of $f$ displaces $\theta$ by $-\eta g_{f}/\sqrt{\hat v}$, and the
displacement decays as $(1-k)^t$. Exposures at a rate of $r$ per step hold it at $-r\,g_f/h$ on time average, a value
independent of $\eta$ and $\hat v$, which set only the relaxation time $1/k$. The stationary variance of $\theta$ is
$\eta^2(\sigma^2+s^2)/\big(\hat v\,(1-(1-k)^2)\big)\approx\eta(\sigma^2+s^2)/(2h\sqrt{\hat v})$. In noise-dominated
coordinates ($\sigma^2\approx v$), error in both moments multiplies it by $\sqrt{1+\rho}$, error only in the second
moment by $1/\sqrt{1+\rho}$, as a learning rate divided by $\sqrt{1+\rho}$ would, and error only in the first moment by
$1+\rho$. A smaller step thus delays the approach to the same time-averaged displacement, whereas update noise widens
the spread around it for as long as the noise lasts. If recall depends on a fact's displacement relative to this
spread, the first effect is a delay that fades once displacements have equilibrated and the second a lasting penalty
that weighs most on facts with small displacements. This link to recall is heuristic.

\section{Parameter-Free Prediction}\label{app:precond}
\begin{table}[tbp]
\centering
\caption{Configurations of Figure~\ref{fig:mech}(c): multiplier predicted by Eq.~\eqref{eq:mf} (geometric mean over 96 facts and 3 checkpoints), observed multiplier at matched tokens with 95\% interval and seed count, and gradient-error bias ratio (about $1/32$ if zero-mean, Appendix~\ref{app:precond}).}
\label{tab:theory}
\small
\begin{tabular}{@{}lrrrr@{}}
\toprule
Configuration & Predicted & Observed [95\% CI] & $n$ & Bias ratio \\
\midrule
Block-shaped gradient noise, 0.5$\times$ & 1.02 & 1.05 [0.86, 1.27] & 3 & 0.03 \\
FP8 & 1.02 & 1.11 [1.00, 1.22] & 6 & 0.30 \\
Forward noise, 0.5$\times$ & 1.03 & 1.08 [0.90, 1.29] & 3 & 0.51 \\
i.i.d.\ gradient noise, 0.5$\times$ & 1.05 & 1.02 [0.95, 1.09] & 3 & 0.03 \\
NVFP4, forward weights only & 1.05 & 1.11 [0.94, 1.31] & 3 & 0.81 \\
NVFP4, backward only & 1.06 & 1.13 [1.04, 1.21] & 6 & 0.03 \\
Block-shaped gradient noise, 1$\times$ & 1.07 & 1.11 [1.03, 1.21] & 6 & 0.03 \\
NVFP4, forward activations only & 1.08 & 1.14 [1.03, 1.26] & 3 & 0.71 \\
Forward noise, 1$\times$ & 1.11 & 1.19 [1.08, 1.31] & 6 & 0.79 \\
i.i.d.\ gradient noise, 1$\times$ & 1.13 & 1.25 [1.09, 1.44] & 6 & 0.03 \\
NVFP4, forward only & 1.15 & 1.18 [1.07, 1.30] & 6 & 0.85 \\
NVFP4, FP8 backward & 1.15 & 1.23 [1.09, 1.39] & 6 & 0.85 \\
NVFP4 & 1.20 & 1.32 [1.16, 1.50] & 6 & 0.80 \\
Block-shaped gradient noise, 2$\times$ & 1.23 & 1.38 [1.24, 1.53] & 3 & 0.03 \\
i.i.d.\ gradient noise, 2$\times$ & 1.36 & 1.33 [1.15, 1.53] & 8 & 0.03 \\
MXFP4 & 1.44 & 1.61 [1.38, 1.88] & 6 & 0.94 \\
Forward noise, 2$\times$ & 1.67 & 1.44 [1.31, 1.58] & 3 & 0.89 \\
\bottomrule
\end{tabular}
\end{table}

For each BF16 checkpoint (40M synthetic, seeds 0--2, final weights) we draw 32 training batches of the original size
($64\times512$ tokens, with gradients accumulated over 8 micro-batches for memory) and compute, per batch, the BF16
gradient $g$ and the gradient $g+e$ of every configuration with its GEMMs on the same batch. We estimate
$v_i$ as the mean of $g_i^2$ and $s_i^2$ as the mean of $e_i^2$ over batches, using all parameters, including the
unquantized embeddings and output head, whose gradients change through the backward pass. For 32 probe facts in each
of the bins seen 4, 8 and 16 times we approximate $g_f$ by the gradient of the summed value-token loss under the probe
prompt,
evaluate Eq.~\eqref{eq:mf} with Adam's $\epsilon=10^{-8}$ added to both square roots, and report the geometric mean
over facts. For the gradient-noise configurations the error has zero mean conditionally on the gradient by
construction (independent Gaussian noise), which also removes the cross term $\mathbb{E}[g_ie_i]$. Backward-only
\nvfp{} comes close to this case. Its stochastic rounding draws independently for every operand of every backward
GEMM, and the backward pass is linear in each rounded operand given the forward activations, so unbiased rounding
would make its error conditionally zero-mean. The emulated rounding is slightly biased, because the E4M3 block scale
is rounded to nearest and, when it is rounded down, the largest elements of a block saturate. In 2{,}000 repeated
roundings of fixed Gaussian, Laplace and Student-$t_3$ tensors, this bias affects about 3\% of elements, carries at
most about 1\% of the error power and slightly shrinks magnitudes ($\mathbb{E}[xb]/\mathbb{E}[x^2]$ between $-0.6\%$
and $-0.3\%$ for input $x$ and bias $b$). For MXFP4 it carries 7--12\% of the error power. The BF16 reference itself
rounds with a relative error of at most $2^{-9}$, against a relative RMS error of about 0.14 for one \nvfp{} rounding.
As a check on the mean error over batches only, the bias ratio $\sum_{i\in I}\bar e_i^2/\sum_{i\in I}s_i^2$, where
$\bar e_i$ is the mean of $e_i$ over the 32 batches and $I$ contains every 64th coordinate, is about $1/32$ for a
zero-mean error. It is 0.03 for every backward-noise configuration and for backward-only \nvfp{}, and 0.3--0.94 for
every configuration whose forward pass is quantized or perturbed, including FP8 and MXFP4 (open markers in
Figure~\ref{fig:mech}(c), which mark ratios above 0.1). Table~\ref{tab:theory} lists the prediction, the observation
and this ratio for every configuration.

\paragraph{Estimator sensitivity.}
The value of Eq.~\eqref{eq:mf}, though not its ranking, depends on how the normalizer is estimated. On the seed-0
checkpoint with 64 batches, $\log m_{\mathrm{pred}}$ is 0.88 times its 32-batch value (median over the 17
configurations, range 0.78--0.96), and replacing $\mathbb{E}[g_i^2]$ by Adam's normalizer, an exponential average with
$\beta_2=0.95$ of squared minibatch gradients whose error comes from an independent batch, multiplies it by 1.15
(1.05--1.40). At 64 batches the two corrections nearly cancel. Through the origin, the slope of
$\log m_{\mathrm{obs}}$ on $\log m_{\mathrm{pred}}$ over all 17 configurations is 1.07 both for the 32-batch estimate
used in the paper and for Adam's normalizer at 64 batches, and Spearman's correlation is 0.96 under every estimator.
The near-unit slope averages configurations that deviate in opposite directions. For the seven configurations with
zero-mean gradient error the two slopes are 1.22 and 1.23, whereas forward noise at twice the error is overestimated
(1.67 predicted against 1.44$\times$ observed).

\section{Split-Moment Analyses and Pre-Specified Tests}\label{app:prereg}
\paragraph{Split-moment variants.}
The split-moment variants inject the i.i.d.\ gradient noise of the noise configuration, at the \nvfp{} error or twice
it. At every step a
second backward pass without noise computes the BF16 gradient of the same batch, and both gradients are clipped with
the factor of the noisy gradient's global norm, so that the two moments differ only by the injected noise. In the
second-moment-only variant Adam's first moment receives the BF16 gradient and its second moment the noisy one, and
the first-moment-only variant does the reverse. Bias correction, decoupled weight decay and the schedule are those of AdamW with
$\beta=(0.9,0.95)$. Seed-0 reruns that log the global gradient norm at every step show that this shared clipping
factor couples the moments only weakly. The median norm is 0.18--0.21 against the threshold of 1, and clipping acts
on 3.3\% of steps for BF16, 3.7\% for noise in both moments, 3.4\% for noise in the second moment only and 3.3\% for
\nvfp{}. It acts on about 10\% of the first 30\% of steps (9.6--10.9\%) and on at most 0.9\% afterwards. In the
second-moment-only run the noisy gradient's norm exceeds the BF16 gradient's by 3.9\% at the median (0.2--6.4\%,
5th--95th percentile over steps) and varies by at most 1.5\% within 20 steps (95th percentile of the coefficient of
variation), so the factor depends on the noise level rather than on the particular draw. Clipping acts on the same
steps on which it would act for the BF16 gradient, and the two factors differ by $2\times10^{-5}$ on average.
The reruns end within 0.0007 of the original runs' final validation loss and within 0.0012 of their mean recall.

\paragraph{Analyses over training and by exposure count.}
For Figure~\ref{fig:split}(a) we recompute the token-matched multiplier of Definition~\ref{def:mult} from the recall
curves of a run and of BF16 at $10^{-3}$ at every evaluation step they share and average its logarithm over seeds.
For Figure~\ref{fig:split}(b), at every evaluation from step 700 on (after the loss plateau) and in every exposure
bin where the recall of both runs lies in [0.2, 0.8], the gap is the bin's $\log_2$ exposure count minus the $\log_2$
exposure count at which the reference's recall curve at the same step reaches the recall of the run. Per seed, the
within-bin slope regresses the gaps on the step with a fixed effect per bin, the between-bin slope regresses them on
$\log_2$ exposures with a fixed effect per step, and the profile averages the gaps of each bin. We report means over
seeds with $t$ intervals, for bins with at least three seeds (six for facts seen twice, eight otherwise). The
update-noise share is $\bar d/\bar f$, where $f$ is the $\log_2$ multiplier of noise in both moments and $d$ its
difference from noise in the second moment only, both against BF16 at $10^{-3}$, averaged over the eight seeds, and
its interval resamples seeds 20{,}000 times.

\paragraph{Failure rule.}
A run fails if its validation loss first reaches $V^*$, the median validation loss of the BF16 runs at $10^{-3}$ near
30\% of training, later than twice their median step or never, or if after 10\% of training it rises more than one nat
above its running minimum. We fixed this rule with the learning-rate compensation test below, before its runs and
after those of E1--E3. Among the latter it flags six of the eight first-moment-only runs at twice the \nvfp{} error,
all by the second criterion, and both runs of seed 4 at $2\times10^{-3}$, which reach $V^*$ only after 90\% of
training.

\paragraph{E1--E3.}
We recorded the following predictions and decision rules in the analysis code before the runs, and the version
history of the code records the order. E3: at $10^{-3}$, noise
in the second moment only matches BF16 at $10^{-3}/A$, with $A=1.356$ from Eq.~\eqref{eq:mf} for this noise
(geometric mean over the three checkpoints) and the product of rate and decay fixed, and the reduced rate costs about
1.09$\times$ against BF16 at $10^{-3}$. Observed: 1.02$\times$ [0.93, 1.13] and 1.08$\times$ [0.96, 1.21] (8 seeds),
as predicted. E2: at $2\times10^{-3}$, above the optimal rate, the same variant was predicted to save exposures against
BF16 at $2\times10^{-3}$ (0.70$\times$ [0.57, 0.86], from a quadratic through BF16's recall at three rates) and to match
BF16 at $2\times10^{-3}/A$. Over the six seeds it gives 0.56$\times$ [0.10, 3.25], outcome B of the rule (point below 1,
interval containing 1), but only because BF16 failed in seed 4. Without that seed it gives 1.11$\times$ [0.92, 1.35],
with no saving. Against BF16 at $2\times10^{-3}/A$ the variant costs 1.66$\times$ [1.33, 2.07], above 1 in all six seeds
(1.58$\times$ [1.24, 2.00] without seed 4), although BF16 at $2\times10^{-3}/A$ does beat BF16 at $2\times10^{-3}$ as
predicted (0.70$\times$ [0.56, 0.89] without seed 4). The uniform-cut reading thus fails above the optimum. A
candidate explanation, which we have not tested, is the per-coordinate structure of the cut. The measured mean of
$v_i/(v_i+s_i^2)$ at this noise is 0.43--0.66 for the embedding and the weight matrices of the blocks but 1.0 for the
output head, whose gradient carries no injected error and keeps the full rate. E1: (i) for i.i.d.\ noise the median $s_i^2/v_i$ of the query and
key rows was predicted to be at least ten times that of the value rows. It is 2.7--3.1 times (below 1 for
block-shaped noise and backward-only \nvfp{}, as predicted). (ii) The slope of $\log m_{\mathrm{obs}}$ on $\log m_{\mathrm{pred}}$
after replacing $\mathbb{E}[g_i^2]$ by Adam's normalizer was to attribute the underestimate of Eq.~\eqref{eq:mf} to
heavy-tailed per-batch gradients, a bias of the estimator (at most 1.10), or to the training dynamics (at least
1.15). The pre-specified statistic, which applies the normalizer correction to the 32-batch estimate, was 1.13 when
first computed, with the forward-noise configurations evaluated with the noise on. With the corrected readout of
Section~\ref{sec:mech} it is 0.99, which the rule attributes to the estimator. Over the seven zero-mean
configurations, which
the readout does not affect, the same statistic is 1.12, between the thresholds, and the underestimate persists under
every estimator (Appendix~\ref{app:precond}).

\paragraph{Learning-rate compensation.}
Before the runs we fixed two primary statistics, both paired by seed and token-matched against BF16 at $10^{-3}$, with
the product of rate and decay at $10^{-4}$: $R_1$, the multiplier of noise at twice the \nvfp{} error with the rate
raised by $A=1.356$ over that at $10^{-3}$ (seeds 0--7), and $R_2$, the same for \nvfp{} with its $A=1.199$ (seeds
0--5). If the cost were only the shrunken step, raising the rate would remove it, giving $R_1\approx0.75$ and
$R_2\approx0.76$, the inverses of the multipliers at $10^{-3}$. If only the step-size cost is removed, $R_1\approx0.91$,
the inverse of the second-moment-only multiplier, or BF16's own multiplier at the raised rate (measured in the same
test and expected near 1.10) if larger updates cost as they do for BF16, and $R_2$ is 0.95--1.05. The rate counts as
restoring the cost if the upper 95\% bound lies below 0.83 for $R_1$ (0.85 for $R_2$), as not restoring it if the
lower bound lies above, and as partial otherwise. Every seed enters the primary analysis, and a secondary analysis
drops seeds with a run that fails the rule above.
Both primary statistics fall in the region where the rate does not restore the cost, with $R_1=1.08$ [0.95, 1.24]
(per seed 0.83--1.29) and $R_2=0.99$ [0.87, 1.13] (0.89--1.20). No run failed the rule, so the secondary analysis equals
the primary one. $R_1$ lies between the two forms of the signal-to-noise account, 0.91 and BF16's own multiplier at
the raised rate, 1.20$\times$ [1.08, 1.35]. The secondary comparisons also follow that account. The noise at the
raised rate costs 1.44$\times$ [1.31, 1.58] against BF16 at $10^{-3}$ (pre-specified range 1.21--1.46 for the
signal-to-noise account and about 1 for the step-size account) and 1.19$\times$ [1.09, 1.30] against BF16 at the
raised rate (1.1--1.33 against about 0.9).

\section{Example Biography and Probes}\label{app:example}
Each exposure of a person is a newly sampled biography, with the six attribute sentences in random order, each from
one of the five templates of its attribute. The box shows one biography of a probed person from the bin seen eight times,
generated with the code and data seed of the 40M runs, and the six probe prompts with their values. A fact counts as
recalled if the model's top prediction is the first value token (bold). In the word-level vocabulary every word
and punctuation mark is one token. In the natural-text regime the same text is inserted as a separate document
between FineWeb-Edu documents and tokenized by the byte-level BPE. Its first value token is then a whole city or
major but is shared by several universities and by several companies (41 distinct first tokens among the probed
universities and 50 among the probed companies). Exact match of the whole value is unaffected
(Appendix~\ref{app:robust}).
\begin{center}
\fbox{\parbox{0.96\textwidth}{\small
\textbf{Biography.} The alma mater of Lunsawi MVorbel LOrlochefen is UKami University . Lunsawi MVorbel LOrlochefen
completed a degree in SDasdri . Lunsawi MVorbel LOrlochefen relocated to CChefen for work . Lunsawi MVorbel
LOrlochefen celebrates a birthday on March 26 , 1950 . Lunsawi MVorbel LOrlochefen was employed by KBelvorpa Inc .
Lunsawi MVorbel LOrlochefen was born in CWirapa .\\[5pt]
\textbf{Probes.}\\[2pt]
\begin{tabular}{@{}l@{\qquad}l@{}}
Lunsawi MVorbel LOrlochefen was born on & \textbf{March} 26 , 1950 \\
Lunsawi MVorbel LOrlochefen was born in & \textbf{CWirapa} \\
Lunsawi MVorbel LOrlochefen studied at & \textbf{UKami} University \\
Lunsawi MVorbel LOrlochefen majored in & \textbf{SDasdri} \\
Lunsawi MVorbel LOrlochefen worked at & \textbf{KBelvorpa} Inc \\
Lunsawi MVorbel LOrlochefen worked in & \textbf{CChefen}
\end{tabular}}}
\end{center}

\end{document}